\documentclass{article}

\PassOptionsToPackage{numbers, compress}{natbib}
\usepackage[preprint]{neurips_2026}

\usepackage[utf8]{inputenc} 
\usepackage[T1]{fontenc}    
\usepackage{hyperref}       
\usepackage{url}            
\usepackage{booktabs}       
\usepackage{amsfonts}       
\usepackage{nicefrac}       
\usepackage{microtype}      
\usepackage{xcolor}         
\usepackage{graphicx}
\usepackage{amsmath}
\usepackage{multirow}
\usepackage{float}
\usepackage{enumitem}
\usepackage{makecell}  
\usepackage{algorithm}      
\usepackage{algorithmic}
\usepackage{tabularx}
\usepackage{rotating}
\usepackage{wrapfig}
\usepackage{subcaption}
\title{Breaking the Group Size Barrier: Parameter-Efficient Group Dance Generation with Chain-of-Dancers}

\author{%
  Jing Xu \quad
  Cunjian Chen\thanks{Corresponding author: \texttt{cunjian.chen@monash.edu}} \quad
  Qiuhong Ke \\
  Department of Data Science and Artificial Intelligence, \\Monash University, Australia
}

\begin{document}

\maketitle

\begin{abstract}
    Group dance generation aims to synthesize coordinated multi-dancer choreography from music, with broad applications in animation and interactive content creation. This task requires modeling dense inter-person dependencies to ensure spatial coordination, while naturally preserving individual dancer identities. Existing approaches model all dancers jointly with end-to-end transformers, which tie the architecture to a fixed group size and  entangle per-dancer identities across frames. We propose \textbf{ChainDance}, a scalable framework that \emph{reformulates} group dance generation as a \textit{Chain-of-Dancers}: a sequential decomposition over per-dancer conditional distributions, allowing a single model to scale across variable group sizes without retraining and naturally preserving per-dancer identity. Built on a frozen single-dancer diffusion backbone, ChainDance introduces two lightweight modules: a \textbf{Role-Aware Text Encoder} (RATE) for per-dancer semantic conditioning, and a \textbf{Group-Aware Motion Encoder} (GAME) that aggregates previously generated dancers via a distance-weighted graph convolutional network, and incorporates a training-free noise optimization procedure at inference time to enforce global spatial coherence. Experiments on AIOZ-GDance demonstrate that ChainDance achieves state-of-the-art motion quality and group coordination while structurally preserving per-dancer identity, with $3$-$4\times$ fewer parameters and requiring $3$-$6\times$ less training time compared to prior approaches.
\end{abstract}

\section{Introduction}
\begin{figure}[t]
  \centering
  \includegraphics[width=\linewidth]{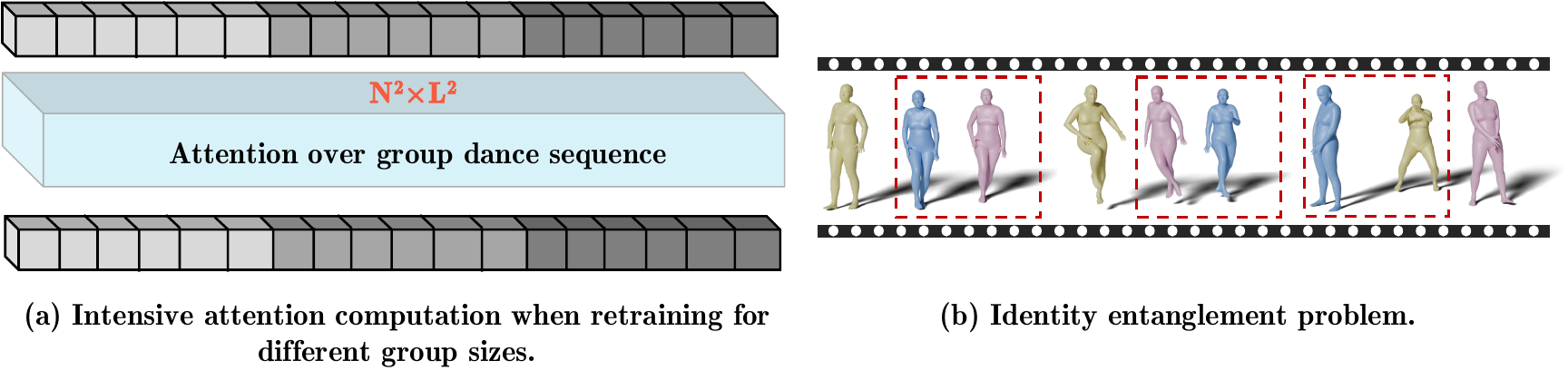}
  \caption{\textbf{Key challenges in group dance generation.} (a) Group-size rigidity: adapting to a new group size $N$ requires retraining from scratch, but self-attention over the concatenated sequence scales as $O((NL)^2)$, making per-size retraining prohibitively expensive. As a result, existing models are locked to a fixed group size. (b) Identity entanglement: joint representations obscure per-dancer identity, causing frequent swaps across frames. We visualize three consecutive frames from CoDancers~\cite{yang2024codancers}: between Frame 1 and Frame 2, the pink and blue dancers abruptly swap positions; in Frame 3, the blue dancer suddenly relocates to the leftmost position. When rendered as video, these inconsistencies appear as visible flickering and identity-switching artifacts.}  
  \label{fig:challenges}
  \vspace{-14pt}
\end{figure}

Group dance generation aims to synthesize coordinated multi-person motion from music while preserving individual identities and interactions~\cite{schwartz1998passacaille,yao2023dance}. Despite recent progress, existing approaches~\cite{le2023controllable,yang2024codancers,dai2025harmonious} are fundamentally limited by how they formulate the problem. Most methods model all dancers jointly within a fixed-size representation, coupling group size to model architecture. This leads to two inherent constraints (Figure~\ref{fig:challenges}): (1) \textbf{group-size rigidity}, requiring retraining to handle different numbers of dancers, and (2) \textbf{identity entanglement}, where individual dancer characteristics are obscured by shared representations.

These limitations are not merely architectural inefficiencies. They stem from a deeper assumption that group choreography must be generated as a single joint process. In contrast, we ask: can group choreography be constructed as a composition of individual motion processes while still capturing inter-person coordination? This question is grounded in real choreographic practice, where professional choreographers first match music-aligned motions to performers, then design spatial formations.

In this work, we answer this question affirmatively by introducing a new formulation of group dance generation as a sequential conditional decomposition over dancers, which we term a \textit{Chain-of-Dancers}. Rather than treating dancers as simultaneously generated tokens, we model each dancer as conditioned on previously generated ones, enabling flexible scaling to arbitrary group sizes while preserving consistent identity structure. This perspective shifts the problem from joint generation to structured composition, decoupling group size from model capacity.

This reformulation enables a single pretrained single-dancer model~\cite{huang2023tg,ghosh2025duetgen,qi2023diffdance} to generalize to multi-dancer scenarios without retraining, while still capturing global coordination through explicit inter-dancer conditioning. Crucially, it allows identity to be preserved by construction, rather than enforced through architectural constraints.

Building on this formulation, we introduce a lightweight instantiation that injects role-specific semantics and inter-dancer interactions into a frozen backbone. We further incorporate a training-free inference procedure~\cite{ota2025pino} to enforce global spatial coherence. Together, these design choices operationalize the proposed decomposition while maintaining parameter efficiency.

Experiments demonstrate that this reformulation achieves strong motion quality and coordination across varying group sizes, while significantly reducing training cost. More importantly, our results suggest that structured decomposition is a viable alternative to joint modeling for multi-agent motion generation, opening a new direction for scalable and controllable choreography synthesis. Our contributions can be summarized as follows:

\begin{itemize}[leftmargin=*, itemsep=1pt, topsep=1pt]
    \item \textbf{Reformulating group dance generation.} We propose a \textit{Chain-of-Dancers} formulation that models group choreography as a sequential conditional composition of individual dancers, shifting from joint generation to structured decomposition and addressing group-size rigidity and identity entanglement.
    \item \textbf{Scalable and identity-preserving generation.} Our approach decouples group size from model architecture, enabling a single pretrained model to generalize to variable numbers of dancers without retraining, while preserving each dancer's identity by construction.
    \item \textbf{Efficient coordination with minimal overhead.} We instantiate the formulation with a Role-Aware Text Encoder, a Group-Aware Motion Encoder, and a training-free inference strategy for global coherence, achieving strong results with significantly reduced training cost.
\end{itemize}

\vspace{-10pt}
\section{Related Work}
\vspace{-5pt}
\subsection{Music-Driven Single Dance Generation}
\vspace{-5pt}
Music-to-dance generation targets rhythmic and stylistically consistent motion synthesis~\cite{joshi2021extensive, siyao2022bailando, siyao2023bailandopp}. Transitioning from rule-based heuristics~\cite{kovar2002motion,kim2003rhythmic,safonova2007construction} to deep learning on large datasets~\cite{lee2019dancing,li2021ai,li2023finedance, siyaoduolando}, various architectures have emerged: RNNs and autoregressive models capture transitions but struggle with error accumulation~\cite{alemi2017groovenet,yalta2019weakly}, while Transformers and GANs/VAEs improve long-range coherence and realism~\cite{sun2020deepdance,ferreira2021learning,lee2019dancing,huang2021choreography,hong2022avatarclip}. Currently, diffusion models define the state-of-the-art via audio-aligned denoising~\cite{tseng2023edge,yin2023dance}. However, most methods focus on music-only conditioning, limiting controllability, semantic expressiveness, and global coherence.

\vspace{-8pt}
\subsection{Group Dance Generation}
\vspace{-5pt}
Group dance generation requires spatial coordination, identity preservation, and semantic diversity across multiple performers~\cite{yao2023dance,tseng2023edge}. Many works adopt a joint modeling strategy that concatenates all dancer motions and processes them with transformer backbones, e.g., GDanceR~\cite{le2023music}, GCD~\cite{le2023controllable}, TCDiff~\cite{dai2025harmonious}, and CoDancers~\cite{yang2024codancers}. While capturing global synchronization, this design fixes the dancer dimension and requires retraining for each group size, blurs per-dancer identity through uniform tokenization, and typically offers only music-only conditioning, widening the semantic gap between music and motion~\cite{wang2025dancechat,lee2019dancing}. These limitations motivate frameworks that scale across group sizes without retraining, support joint music–text control, and explicitly preserve per-dancer identity. A complementary line of work models pairwise leader-follower interactions~\cite{siyaoduolando} but is restricted to fixed two-person generation. In contrast, we generalize this paradigm to variable group sizes via a shared bidirectional pair-wise model with graph-based group conditioning.

\vspace{-6pt}
\subsection{Multimodal Conditioning with Text and Music}
\vspace{-2pt}
Controlling human motion with semantic and rhythmic cues has been widely studied~\cite{wang2025dancechat,chen2020talking}.
Text-to-motion methods map language to coherent movements~\cite{yatim2024space,zeng2025light}, while music-conditioned models learn beat- and phase-aligned trajectories~\cite{zhuang2022music2dance,li2022danceformer,li2021ai}.
Recent works fuse text and audio via cross-modal encoders to couple semantic intent with temporal structure~\cite{yang2025unimumo,gong2023tm2d}, improving controllability over unimodal conditioning. However, simultaneous text-and-music control for single dance is still limited. Most models use music alone or coarse style tags rather than free-form language~\cite{tseng2023edge,huang2020dance}. The gap widens in \emph{group dance}, where role-aware conditioning is rare due to scarce multi-person annotations~\cite{zhou2024avatargpt}. Moreover, many multimodal pipelines emphasize global synchronization or average style, causing identity permutation and weak role differentiation in group dance settings. We address this gap by enabling coordinated group generation under \emph{both} text and music while explicitly maintaining dancer-specific identity and role consistency.

\vspace{-9pt}
\section{Method}
\label{sec:method}
\vspace{-6pt}
Given an input music clip $M$ and an optional dance instruction $T$ (style, dancer count, interaction type), our goal is to synthesize a spatiotemporal group dance $\{\mathbf{X}^{(i)}_{1:L}\}_{i=1}^{N}$, where $\mathbf{X}^{(i)}_{l}$ denotes the motion of dancer $i$ at time $l$.

Rather than modeling all dancers jointly, we introduce a \textit{Chain-of-Dancers} formulation, where group choreography is constructed as a sequence of conditional single-dancer generations. This shifts group generation from joint modeling to structured composition, enabling scalable and identity-consistent synthesis.

This formulation introduces three key challenges: (a) how to inject inter-dancer dependencies into a single-dancer backbone, (b) how to support variable group sizes without retraining, and (c) how to enforce global coordination beyond local interactions.

\begin{figure*}[t]
  \centering
  \includegraphics[width=\textwidth]{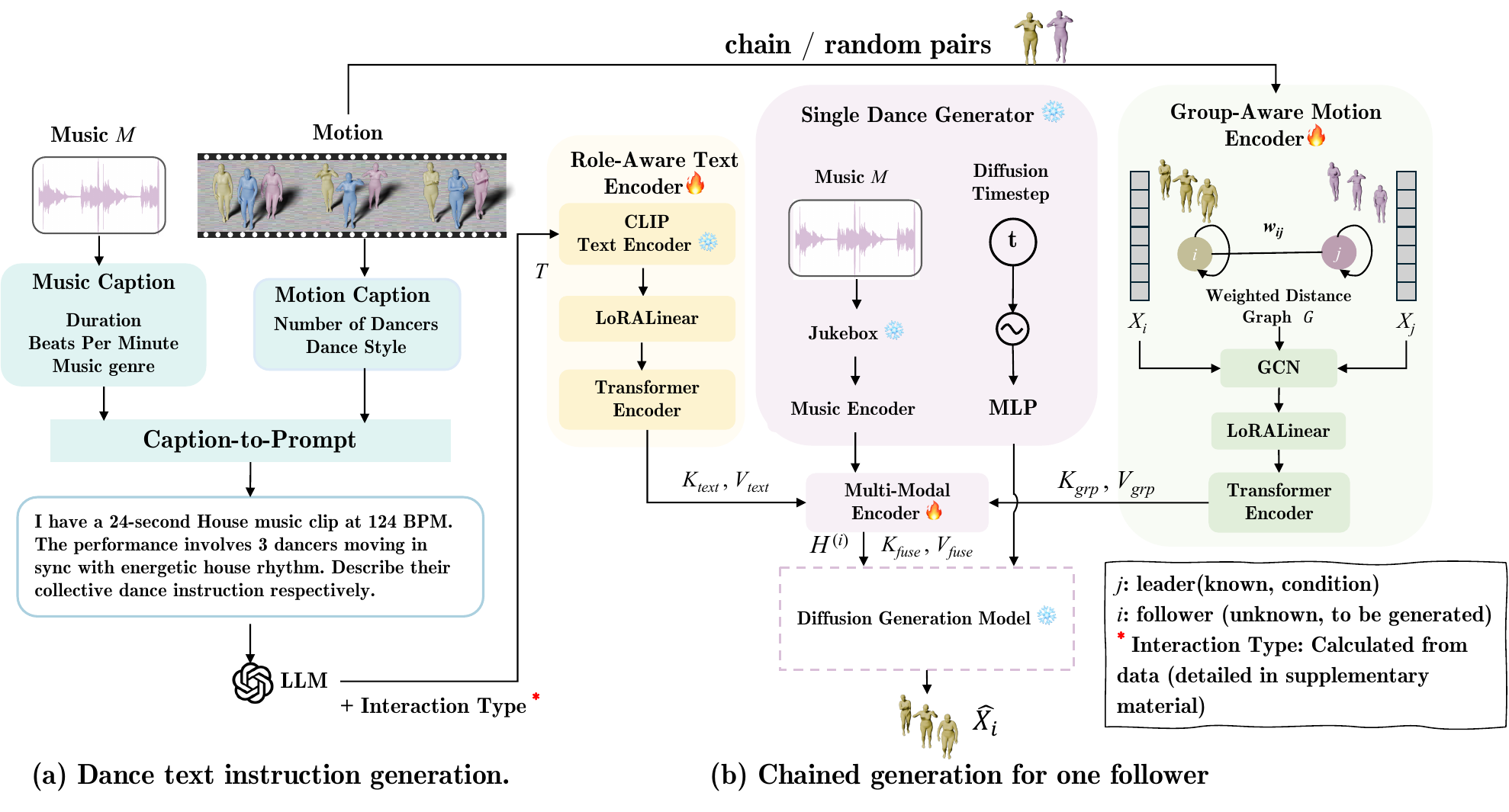}
  \caption{\textbf{Overall framework of ChainDance.} The model builds on a frozen single dance diffusion backbone and introduces lightweight multimodal encoders for text, music, and other dancers. Through sequential chain generation and inference-time global optimization, ChainDance produces coherent and coordinated group choreographies.}
  \label{fig:framework}
  \vspace{-16pt}
\end{figure*}

We address these challenges through \textbf{ChainDance} (Figure~\ref{fig:framework}), a unified framework that combines conditional generation (\S\ref{sec:textprompt}), pairwise decomposition (\S\ref{sec:chaingen}), and inference-time optimization (\S\ref{sec:globalopt}).

\begin{figure}[b]
  \centering
  \vspace{-10pt}
  \includegraphics[width=\linewidth]{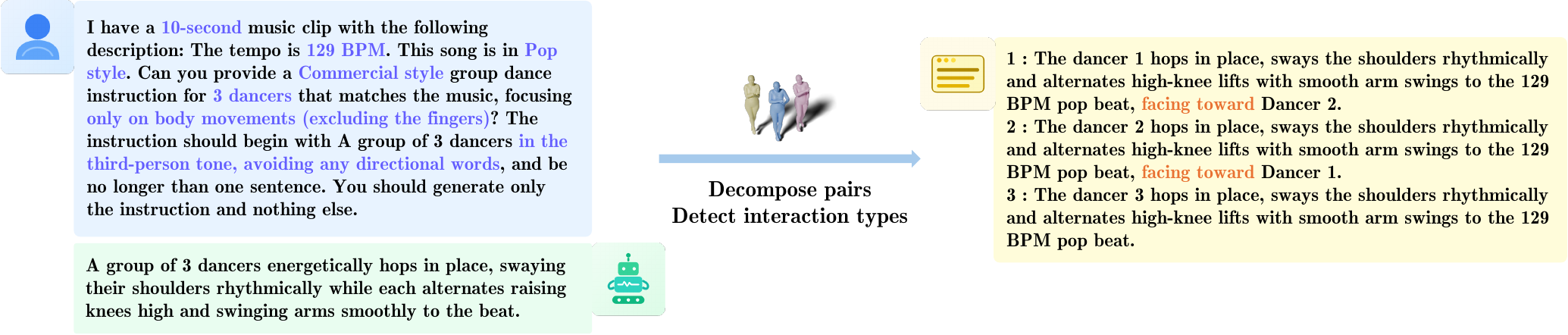}
  \caption{\textbf{Prompt generation and role-specific conditioning.}
  The system converts structured multimodal captions into individualized
  text prompts for each dancer. Based on a global choreography description,
  dancer pairs are randomly sampled and assigned interaction keywords 
  (e.g., \textit{approach}, \textit{keep distance}, \textit{face to}, 
  \textit{lead follow}, \textit{swap}, or \textit{passby}),
  providing diverse yet coherent textual conditions for ChainDance.}
  \label{fig:textprompt}
  \vspace{-12pt}
\end{figure}

\vspace{-8pt}
\subsection{Multimodal Single-Dancer Generation and Inter-Dancer Interaction Modeling}
\label{sec:textprompt}
\vspace{-6pt}
To realize the proposed decomposition, each dancer must be generated conditionally on both global instructions and previously generated dancers. We achieve this by augmenting a pretrained single-dancer diffusion model with two types of conditioning: (a) semantic conditioning from text (roles, interactions, style), and (b) relational conditioning from previously generated dancers.

Rather than modifying the backbone architecture, we inject these signals through lightweight conditioning pathways, allowing the model to remain fixed while adapting to group scenarios. This design ensures that each dancer retains individual identity, and coordination emerges through explicit conditioning, not shared latent entanglement.

\vspace{-4pt}
\noindent\textbf{Textual Instruction Generation.}~As shown in Figure~\ref{fig:textprompt}, we first merge structured multimodal metadata (music captions such as tempo and genre, motion captions such as style and interaction type) into a global choreography prompt. 
We then use an LLM (ChatGPT-4o) to generate a textual dance instruction $T$ that aligns with the music $M$.
This global instruction is decomposed into individualized instructions based on the pairing strategy.
Each dancer pair is assigned an interaction keyword (e.g., \textit{approach}, \textit{keep distance}, \textit{face to}, \textit{lead follow}, \textit{swap}, \textit{passby}) to encourage diverse yet coherent dancer-aware behaviors. 
Existing group dance datasets provide high-quality kinematic trajectories but lack fine-grained textual annotations of inter-dancer interactions. We bridge this gap by automatically deriving interaction keywords from motion samples via geometrically-grounded rule-based detection (Appendix~\ref{sec:rule_interaction}), serving as approximate yet effective weak supervision (Table~\ref{tab:ablation}, w/o RATE).

\vspace{-2pt}
\noindent\textbf{Lightweight Multimodal Encoders.}~While the decomposition reduces complexity, coordination still requires modeling interactions between dancers. We capture these interactions through two complementary mechanisms: (a) \emph{textual interaction cues}, which specify roles and interaction types (e.g., approach, follow, keep distance), and (b) \emph{motion-based relational context}, derived from previously generated dancers. These signals provide structured guidance for coordination, and importantly, interaction modeling is explicit and compositional, rather than implicitly entangled in a joint representation.

To realize these two mechanisms, two parameter-efficient encoders are inserted into the cross-attention layers of the frozen single-dancer diffusion backbone $\mathcal{B}$. Pretrained single-dancer models condition only on a global instruction, whereas group choreography requires each dancer to carry role-specific information (e.g., who leads, who interacts). To bridge this gap, the \textit{Role-Aware Text Encoder} (RATE) encodes dancer-aware instructions using a CLIP text encoder~\cite{radford2021learning}, projects them through a low-rank LoRA module and a lightweight Transformer, and provides them as ($K_{text}$, $V_{text}$) in a cross-attention computation:
\begin{equation}
\mathrm{Attn}\big(\mathbf{H}^{(i)}\big) = \mathrm{softmax}\left(\frac{\mathbf{H}^{(i)} K_{text}^\top}{\sqrt{d}}\right)V_{text},
\end{equation}
where $\mathbf{H}^{(i)}$ denotes the hidden state of dancer $i$.
In this way, style and role information are injected into text embeddings, which are then incorporated into the encoder output (Figure~\ref{fig:framework}) to enable fine-grained semantic control.

The \textit{Group-Aware Motion Encoder} (GAME) conditions on previously generated dancers by constructing a distance-weighted graph $G$ over dancers, where edge weight $w_{ij}$ encodes spatial proximity or interaction strength. A GCN~\cite{kipf2016semi, nie2021gcn, ding2022causal} aggregates neighbor features, refined by a Transformer with LoRA updates, to produce motion-aware ($K_{grp}$, $V_{grp}$) fused with $\mathbf{H}^{(i)}$ through cross-attention (see Appendix~\ref{sec:graph_construction} and \ref{sec:alg_details}). This captures \emph{local} pairwise dependencies. The Role-Aware Text Encoder provides semantic cues via ($K_{text}$, $V_{text}$) from textual prompts, while the Group-Aware Motion Encoder introduces spatial context through ($K_{grp}$, $V_{grp}$) from previously generated dancer $X_j$. 

All these $(K,V)$ pairs are fed into the multimodal encoders together with the music condition. The encoder outputs are fused (through concatenation and projection) into key-value pairs ($K_{fuse}$, $V_{fuse}$) that are applied to the cross-attention layers in the frozen generation model.
Through this mechanism, the hidden state of dancer $i$ integrates textual semantics, music, and prior motions for coherent and coordinated choreography.
Together, these encoders transform the single-dancer diffusion model into a multimodal generator capable of synchronized and identity-consistent group choreography.

\begin{figure*}[t]
  \centering
  \includegraphics[width=\textwidth]{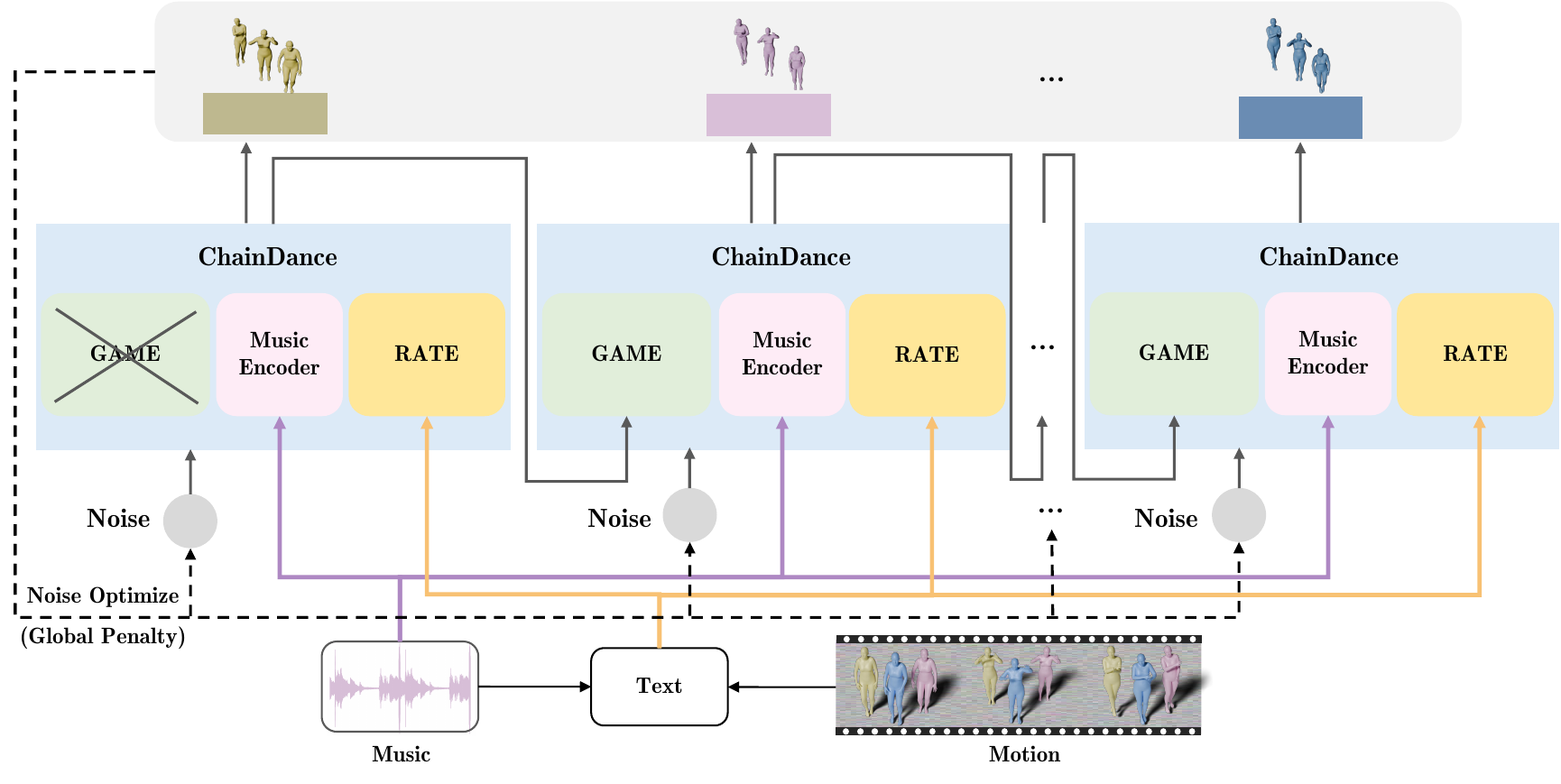}  
  \caption{\textbf{Chain-of-Dancers generation design.} At inference, ChainDance generates dancers sequentially and refines global consistency via diffusion-based optimization. Each dancer is conditioned on the previously generated one, and this process repeats until the whole group is complete. After that, optimize the shared initial noise via gradient descent until the global penalty $L$ converges below the target threshold. All pairs share the weights of the text and motion encoders.}
  \label{fig:flow}
  \vspace{-8pt}
\end{figure*}

\vspace{-5pt}
\subsection{Pair-wise Decomposition for Scalable Generation}
\label{sec:chaingen}
\vspace{-4pt}

A key consequence of the Chain-of-Dancers formulation is that modeling full joint dependencies is unnecessary. Instead, we approximate group choreography through pairwise conditional generation. Concretely, we decompose the generation process into leader–follower pairs, where each dancer is conditioned on another dancer. A single shared model is trained on such pairs and reused across all positions in the chain. At inference time, group choreography is constructed sequentially: each subsequent dancer is conditioned on previously generated ones, and the same model is reused for all steps.
This enables: (a) scalability to varying group sizes and 
(b) no retraining when adding new dancers.

\noindent\textbf{Pair-wise training.}~Given a group of $N$ dancers, we decompose the joint distribution into pairwise conditionals via the chain rule (see Appendix~\ref{sec:assumptions}):
\begin{equation}
\small
P(X_1, \ldots, X_N) = P(X_1) \prod_{i=2}^{N} P(X_i \mid X_{1:i-1}).
\end{equation}
We approximate each conditional $P(X_i \mid X_{1:i-1})$ with a shared pair-wise model: each training sample is a pair $(X_j, X_i)$, where $X_j$ is the conditioning leader and $X_i$ is the follower to be denoised. The same RATE and GAME modules are shared across all pairs, so adding a new dancer at inference adds no learnable parameters.

To avoid directional bias, we adopt \textbf{bidirectional pairing}, allowing the model to learn symmetric interactions rather than fixed roles: both $(i,j)$ and $(j,i)$ samples are included, allowing the model to learn mutual conditioning rather than a fixed leader–follower direction. Two pair sampling strategies are explored:
\vspace{-5pt}
\begin{enumerate}[label=\roman*., leftmargin=*, itemsep=1pt, topsep=2pt]
    \item \textbf{Chain pairing}: pairs follow consecutive dataset indices, e.g., $(D_1, D_2), (D_2, D_3)$. Indices reflect data order, not spatial position.
    \item \textbf{Random pairing}: two dancers are uniformly sampled from the same group.
\end{enumerate}
Both yield comparable results in our experiments, and we use chain pairing as the default. The training loss combines a reconstruction term with four auxiliaries:
\begin{equation}
\small
\mathcal{L}_{\mathrm{local}}
= \mathcal{L}_{\mathrm{simple}}
+ \lambda_{\mathrm{pos}}\mathcal{L}_{\mathrm{pos}}
+ \lambda_{\mathrm{vel}}\mathcal{L}_{\mathrm{vel}}
+ \lambda_{\mathrm{contact}}\mathcal{L}_{\mathrm{contact}}
+ \lambda_{\mathrm{align}}\mathcal{L}_{\mathrm{align}}.
\label{eq:local}
\end{equation}
$\mathcal{L}_{\mathrm{simple}}$ is the reconstruction loss; $\mathcal{L}_{\mathrm{pos}}$, $\mathcal{L}_{\mathrm{vel}}$, $\mathcal{L}_{\mathrm{contact}}$ ensure pose accuracy, smooth transitions, and physical foot–ground contact; $\mathcal{L}_{\mathrm{align}}$ aligns text–music–motion embeddings via cosine similarity. Only RATE and GAME are trainable; the diffusion backbone $\mathcal{B}$ remains frozen.

\noindent\textbf{Inference: from pair to chain.}~At test time, the same model generates groups of variable size $N$ through sequential conditioning (Figure~\ref{fig:flow}). Starting from $X_1$ produced by the frozen single-dancer backbone, we iteratively generate dancer $X_i$ by feeding the most recent leader $X_j$ into GAME, which constructs a distance-weighted graph over previously generated dancers to compute spatial weights, then aggregates this context onto the leader feature to produce $(K_{grp}, V_{grp})$; RATE simultaneously injects the role-specific text prompt for dancer $i$. The pair-wise model naturally generalizes to chained generation, since each step reuses the same shared encoders and the group-aware motion graph grows incrementally as new dancers are added (see Appendix~\ref{sec:graph_construction}). Algorithm details see Appendix~\ref{sec:alg_details}.

\vspace{-5pt}
\subsection{Global Consistency via Inference-Time Optimization}
\label{sec:globalopt}
\vspace{-5pt}
Sequential generation captures local dependencies but does not guarantee global coherence, such as formation structure or collision avoidance among non-adjacent dancers. Instead of increasing model complexity, we enforce global consistency through a training-free inference-time optimization. Specifically, we refine the shared latent variables to minimize a global objective encoding spatial formation constraints, inter-dancer distance relationships, and collision avoidance. This optimization complements the sequential generation process, enabling global coordination without retraining or architectural changes.

Following the principle of PINO~\cite{ota2025pino}, we keep all network parameters frozen and instead  optimize the shared diffusion noise via gradient descent, backpropagating a global penalty through the diffusion sampling process. Crucially, because all dancers are generated from a shared latent variable, this optimization provides a unified control signal that simultaneously adjusts all dancers, enabling group-level coordination that cannot be achieved by per-dancer generation alone (see Table~\ref{tab:pino_compare}). The test-time objective decomposes into global controls and overlap penalties,
\begin{equation}
\small
L \;=\; L_{\mathrm{global}} \;+\; L_{\mathrm{overlap}},
\label{eq:tt_obj}
\end{equation}
where $L_\mathrm{global}$ encodes formation and orientation constraints, and $L_\mathrm{overlap}$ discourages inter-person penetration. The global part is a weighted sum of differentiable controls:

\begin{equation}
\small
L_{\mathrm{global}}
= \sum_{k\in\mathcal{S}} \lambda_k\,L_k,
\label{eq:global_controls}
\end{equation}
where $\mathcal{S}$  is the subset of enabled controls (e.g., root position, movement region, orientation, relative distance), chosen to reflect group choreography constraints.
The overlap term is implemented as a hinge on pairwise root distances:
\begin{equation}
\small
L_{\mathrm{overlap}}
= \frac{1}{T}\sum_{t}\sum_{i<j}\max\!\Bigl(0,\;\delta-\bigl\|p_{\text{root}}^{(i)}(t)-p_{\text{root}}^{(j)}(t)\bigr\|_2 \Bigr).
\label{eq:overlap_hinge}
\end{equation}
At inference, we update only the shared latent to reduce Eqn.~\eqref{eq:tt_obj}, which adds a small latency overhead while improving interaction quality.

\vspace{-8pt}
\section{Experimental Results}
\label{sec:exp}
\vspace{-5pt}
We evaluate ChainDance in standard single- and group-dance settings. Unless stated otherwise, we freeze the backbone and train only RATE and GAME. We report quality, physical plausibility, group coherence, and efficiency, along with ablation studies.
\vspace{-8pt}
\subsection{Experimental Setup}
\label{subsec:setup}
\vspace{-4pt}
\noindent\textbf{Baseline Methods.}~We compare ChainDance with other group dance methods that generate all dancers jointly as high-capacity baselines: \textbf{GCD}~\cite{le2023controllable}, which integrates group-based contrastive learning with diffusion; \textbf{CoDancers}~\cite{yang2024codancers}, which employs music-aligned choreographic units; \textbf{TCDiff}~\cite{dai2025harmonious}, which incorporates trajectory control for temporal modeling; and \textbf{ST\text{-}GDance}~\cite{xu2025stgdance}, which factorizes temporal and spatial modeling to preserve layouts while avoiding parameter blow-up at scale. All methods share the same music inputs, beat extraction, and evaluation protocols. Each baseline is retrained per group size, while our model is trained once and generalizes to varying group sizes.

\vspace{-2pt}
\noindent\textbf{Datasets.}~We pretrain our single-dancer backbones on two standard single dance datasets: AIST++~\cite{li2021ai} and FineDance~\cite{li2023finedance}. AIST++ contains 1,408 paired 3D music–dance sequences across 10 genres, performed by 60 professional dancers. FineDance builds upon AIST++ with more diverse styles and longer clips, offering broader coverage of musical and motion patterns. All evaluations are conducted on AIOZ-GDance~\cite{le2023music}, the only publicly available large-scale group dance benchmark comprising 16.7 hours of music-aligned 3D motion data from over 4,000 dancers, spanning 7 dance styles and 16 music genres. We follow the official train/test split protocol from~\cite{le2023music}. 

\vspace{-2pt}
\noindent\textbf{Evaluation Metrics.}~Following standard practice~\cite{dai2025harmonious,xu2025stgdance}, we evaluate both single- and group-level performance, along with model efficiency:

\vspace{-7pt}
\begin{itemize}[leftmargin=*]
    \item \textbf{Single dance metrics.} 
    Frechet Inception Distance (FID,~$\downarrow$) assesses distribution fidelity, Generation Diversity (Div,~$\uparrow$) quantifies sample diversity, and Physical Foot Contact (PFC, $\downarrow$) measures foot sliding artifacts.
    \vspace{-3pt}
    \item \textbf{Group dance metrics.} 
    Group Motion Realism (GMR,~$\downarrow$) measures feature similarity via Frechet Inception Distance;
    Group Motion Correlation (GMC,~$\uparrow$) evaluates coherence through cross-correlation between generated dancers;
    Trajectory Intersection Frequency (TIF,~$\downarrow$) counts the frequency of inter-person body-part collisions.
    
    \item \textbf{Efficiency.} 
    We report FLOPs, parameter count, training time per epoch, and inference time.
\end{itemize}
\vspace{-6pt}
\noindent\textbf{Training Strategy.}~We freeze the pretrained backbone and train the two lightweight LoRA encoders. We adopt a \emph{teacher forcing} strategy during training, where each follower conditions on the ground-truth motion of previously generated dancers, perturbed with noise to reduce the train-inference gap. At inference, the model operates auto-regressively, conditioning on its own predictions. We optimize using the Adam optimizer with cosine learning rate decay.

\vspace{-3pt}
\noindent\textbf{Implementation Details.}~We train our model on a dual RTX 4090 GPU setup. ChainDance uses the Lodge~\cite{li2024lodge} backbone pretrained on FineDance and employs a chain pairing strategy with bidirectional pairs during training for the main results. Following previous works~\cite{dai2025harmonious,wang2025dancechat}, the loss function includes weighted terms for position ($\lambda_\mathrm{pos}=0.646$), velocity ($\lambda_\mathrm{vel}=2.964$), and contact ($\lambda_\mathrm{contact}=10.942$) to model motion dynamics, along with a cross-modal alignment loss ($\lambda_\mathrm{align}=1.0$) to ensure consistency between the generated motion and the input text/music. For all group sizes, the sequence length is set to 120 frames. Our model was trained for 1000 epochs using the Adam optimizer with an initial learning rate of 0.001 and a cosine learning rate scheduler. At inference, we adopt a conservative KV cache that reuses final-projection key–value states across dancers and maintains a causal temporal cache to avoid recomputing past frames. Code will be released upon acceptance.

\begin{table*}[b]
\centering
\vspace{-12pt}
\caption{\textbf{Single and Group dance performance on AIOZ-GDance.} Best results are \textbf{bold}; second-best are \underline{underlined}. * denotes models without the global noise optimization penalty.}
\label{tab:groupdance}
\setlength{\tabcolsep}{4.5pt}
\renewcommand{\arraystretch}{1.12}
\resizebox{\textwidth}{!}{
\begin{tabular}{l|ccc|ccc|ccc|ccc}
\hline
\multirow{2}{*}{Method} &
\multicolumn{3}{c|}{\textbf{1 Dancer}} &
\multicolumn{3}{c|}{\textbf{2 Dancers}} &
\multicolumn{3}{c|}{\textbf{3 Dancers}} &
\multicolumn{3}{c}{\textbf{4 Dancers}} \\
\cline{2-13}
& FID$\downarrow$ & Div$\uparrow$ & PFC$\downarrow$
& GMR$\downarrow$ & GMC$\uparrow$ & TIF$\downarrow$
& GMR$\downarrow$ & GMC$\uparrow$ & TIF$\downarrow$
& GMR$\downarrow$ & GMC$\uparrow$ & TIF$\downarrow$ \\
\hline
GCD~\cite{le2023controllable}
& 39.24 & 9.64 & 2.53
& 34.39 & 80.32  & 0.17
& 30.22 & 80.22 & 0.19
& 36.28 & 81.82 & 0.13 \\
CoDancers~\cite{yang2024codancers}
& \underline{23.98} & 9.48 & 3.53
& 24.55 & 72.52 & \textbf{0.08}
& 26.34 & 74.22 & \textbf{0.08}
& 26.44 & 75.34 & \textbf{0.10} \\
TCDiff~\cite{dai2025harmonious}
& 37.31 & \textbf{14.01} & \underline{0.51}
& 15.77 & \textbf{81.92} & 0.12
& 15.36 & \underline{82.77} & 0.11
& 13.44 & 81.70 & 0.15 \\
ST\mbox{-}GDance~\cite{xu2025stgdance}
& 28.87 & 12.82 & 0.97
& 19.42 & 80.52 & 0.12
& 14.76 & 81.24 & 0.11
& 14.02 & 81.42 & \textbf{0.10} \\
\hline
ChainDance* 
& \textbf{23.12} & \underline{13.85} & \textbf{0.50}
& \underline{14.10} & \underline{81.52} & \underline{0.11}
& \underline{12.44} & 82.32 & 0.11
& \underline{13.28} & \underline{82.13} & \underline{0.12} \\
ChainDance
& \textbf{23.12} & \underline{13.85} & \textbf{0.50}
& \textbf{13.67} & 81.28 & 0.12
& \textbf{11.63} & \textbf{83.74} & \underline{0.10}
& \textbf{12.59} & \textbf{82.16} & 0.13 \\
\hline
\end{tabular}
}
\vspace{-10pt}
\end{table*}

\vspace{-9pt}
\subsection{Quantitative Comparisons Against Baseline Models}
\label{subsec:mainresults}
\vspace{-2pt}
\noindent\textbf{Single Dance (1 Dancer).}~As shown in Table~\ref{tab:groupdance}, our model achieves the best overall performance in single-dancer generation, reaching the lowest FID (23.12) and PFC (0.50) and the second-highest diversity (13.85). This indicates superior motion quality and stronger music alignment compared to ST-GDance~\cite{xu2025stgdance} and TCDiff~\cite{dai2025harmonious}.
These results also confirm that the proposed multimodal encoders effectively enhance motion quality and controllability under text–music supervision, without compromising the fidelity of the pretrained single-dancer backbone. 

\vspace{-2pt}
\noindent\textbf{Group Dance (2–4 Dancers).}~For multi-dancer choreography, ChainDance achieves the best GMR$\downarrow$, indicating the highest motion realism through improved feature distribution matching. It also achieves the best GMC$\uparrow$ at three and four dancers, demonstrating well-balanced group dance quality.
Specifically, for three dancers, our model reduces GMR to \textbf{11.63} and achieves the highest GMC (\textbf{83.74}), indicating superior coordination and spatial harmony. 
Even as the number of dancers increases to four, ChainDance maintains more coherent and rhythmic stability, outperforming all baselines with lower GMR and higher GMC values. Moreover, the variant without the global optimization penalty still outperforms other approaches on most metrics, demonstrating that the core chain-of-dancers design and pairwise training strategy are sufficient to ensure strong inter-dancer coordination. 
These results validate the model’s ability to generalize to variable group sizes, supporting efficient and coherent choreography generation from single to multi-person scenarios. To mitigate data confounds, Appendix~\ref{otherabla} reports ChainDance pretrained solely on AIOZ-GDance, which remains competitive (Table~\ref{tab:otherdataset}). Visualizations for larger groups are available on the supplementary webpage.

\begin{wraptable}{r}{0.5\linewidth}
\centering
\vspace{-6pt}
\caption{\textbf{Efficiency comparison (3 dancers group).} Best results are \textbf{bold}; second-best are \underline{underlined}. * denotes models without the global noise optimization penalty.}
\label{tab:efficiency_group}
\setlength{\tabcolsep}{4.5pt}
\footnotesize                                    
\renewcommand{\arraystretch}{1.12}
\resizebox{\linewidth}{!}{%
\begin{tabular}{l|cccc}
\hline
Method &
\makecell{FLOPs\\(G)$\downarrow$} &
\makecell{Params\\(M)$\downarrow$} &
\makecell{Train/epoch\\(min:sec)$\downarrow$} &
\makecell{Inf Time\\(min:sec)$\downarrow$} \\
\hline
GCD~\cite{le2023controllable}          & 27.69 & 62.16 & 1{:}04 & 0{:}05 \\
Codancers~\cite{yang2024codancers}     & 58.95 & 59.32 & 0{:}42 & \textbf{0{:}02} \\
TCDiff~\cite{dai2025harmonious}        & 22.04 & 62.48 & 1{:}29 & 0{:}08 \\
ST\mbox{-}GDance~\cite{xu2025stgdance} & \textbf{6.75} & \underline{50.21} & \underline{0{:}37} & \underline{0{:}03} \\
\hline
ChainDance* & \underline{16.92} & \textbf{15.27} & \textbf{0{:}14} & 0{:}06 \\
ChainDance  & \underline{16.92} & \textbf{15.27} & \textbf{0{:}14} & 0{:}15 \\
\hline
\end{tabular}%
}
\vspace{-6pt}
\end{wraptable}
\vspace{-4pt}
\noindent\textbf{Computational Efficiency.}~In Table~\ref{tab:efficiency_group}, our model remains compact and trainable, evidenced by the second lowest FLOPs (16.92G), the fewest parameters (15.27M), and the fastest training time per epoch (0:14). Without the optional test-time refinement, inference reduces from 0:15 to 0:06, comparable to the fastest baselines while preserving generation quality. This overhead is adjustable (fewer optimization steps trade a small amount of accuracy for speed; see appendix) and requires no retraining. In summary, ChainDance has a small model size, fast training, and produces high-quality group dances with controllable inference.

\vspace{-4pt}
\noindent\textbf{Controllability and User Study.}~We report quantitative controllability evaluation across 10 music genres in Appendix~\ref{sec:controllability} and a 30-participant user study in Appendix~\ref{sec:user_study}.

\vspace{-8pt}
\subsection{Qualitative Results}

\begin{figure}[H]
  \centering
  \vspace{-10pt}
  \includegraphics[width=0.83\textwidth]{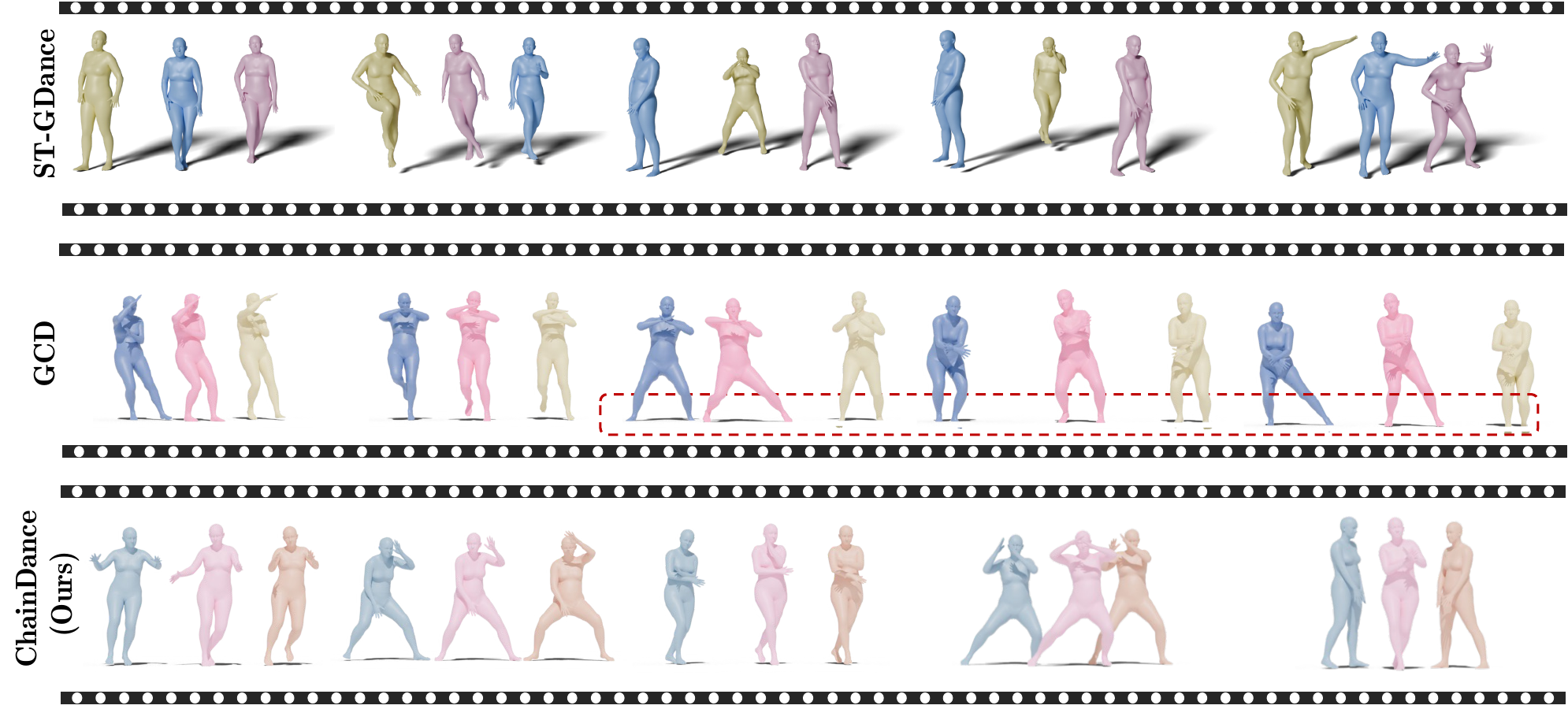}
  \caption{\textbf{Qualitative results of generated 3-dancers group sequence compared with baselines.}}
  \vspace{-12pt}
  \label{fig:viz}
\end{figure}

Figure~\ref{fig:viz} compares dance sequences generated by our method, GCD, and ST-GDance under the same clip length and frame stride. ST\mbox{-}GDance exhibits frequent \emph{apparent} ``position swaps'' between dancers, which do not reflect true choreographic exchanges but arise from the \emph{identity-shift} issue, where global motion remains plausible while per-dancer temporal continuity is lost. Meanwhile, GCD shows interpenetration artifacts (red boxes), where the rightmost dancer's foot penetrates the ground plane, indicating weak physical plausibility. In contrast, ChainDance maintains stable identities and consistent spatial relations, producing collision-free trajectories and realistic ground contact throughout the sequence. This directly supports the identity preservation claim: while joint modeling architecturally permits dancer-feature swaps, ChainDance's sequential decomposition structurally precludes them, since each dancer is an independent conditional sample with no shared feature dimensions. More qualitative videos are on the supplementary webpage.

\vspace{-12pt}
\subsection{Ablation Study}
\label{subsec:ablation}
\vspace{-5pt}
\noindent\textbf{Ablation Settings.}~To better understand the contribution of each component in our framework, we conduct ablation studies on the group dance generation task using the Lodge backbone pretrained on FineDance. We progressively remove or disable key modules to isolate their impact.

\vspace{-3pt}
\emph{w/o GAME}: The follower dancers are generated independently without conditioning on the motion of previously generated dancers. This removes cross-person interaction modeling.

\vspace{-2pt}
\emph{w/o RATE}: The system operates without semantic dance-style guidance, relying solely on music and motion context.

\vspace{-2pt}
\emph{w/o Global Penalty}: We remove the global motion-level consistency penalty at inference, which is designed to prevent inter-person collisions and encourage coherent global formation.

The sequence length is set to 120 frames. All models are trained and evaluated following the same protocol, and group-level metrics are computed on group dance sequences from AIOZ-GDance.

\begin{wraptable}{r}{0.5\linewidth}    
  \centering
  \vspace{-12pt}                       
  \caption{\textbf{Ablation study for group dance generation (3 dancers).} Results are obtained on the Lodge backbone pretrained on FineDance. \textbf{Bold} indicates the best result.}
  \footnotesize
  \begin{tabular}{lccc}
    \toprule
    Method & GMR$\downarrow$ & GMC$\uparrow$ & TIF$\downarrow$ \\
    \midrule
    w/o GAME & 17.46 & 79.14 & 0.19 \\
    w/o RATE                & 14.32 & 81.98 & 0.11 \\
    w/o Global Penalty      & 12.44 & 82.32 & 0.11 \\ 
    \midrule
    \textbf{ChainDance}     & \textbf{11.63} & \textbf{83.74} & \textbf{0.10} \\
    \bottomrule
  \end{tabular}
  \label{tab:ablation}
\end{wraptable}

\noindent\textbf{Results.}~~As summarized in Table~\ref{tab:ablation}, removing the group-aware motion encoder (w/o GAME) leads to the most pronounced degradation (GMR 17.46, GMC 79.14), highlighting the importance of modeling motion dependencies among dancers. Disabling the role-aware text prompt (w/o RATE) substantially degrades both GMR (from 11.63 to 14.32) and GMC (from 83.74 to 81.98), suggesting textual guidance improves both stylistic synchronization and spatial coordination; beyond raw metrics, RATE also provides a human-interpretable interface for per-dancer control (Appendix~\ref{sec:controllability}). Removing the global penalty (\emph{w/o Global Penalty}) leads to moderate degradation across all metrics, confirming its role in inter-person coherence. The full model achieves the best GMR (11.63), GMC (83.74), and TIF (0.10), demonstrating complementary component effects.

\vspace{-8pt}
\subsection{Different Backbones and Pretraining Datasets}

\begin{table}[H]
\centering
\vspace{-16pt}
\caption{\textbf{Comparison across backbones and pretraining datasets.} Best results are \textbf{bold}, and second-best are \underline{underlined}.}
\label{tab:backbone_dataset}
\footnotesize
\setlength{\tabcolsep}{3pt}
\renewcommand{\arraystretch}{1.1}
\resizebox{0.95\textwidth}{!}{
\begin{tabular}{l|cccccc|cccccc}
\toprule
\multirow{3}{*}{Method} 
& \multicolumn{6}{c|}{w/o Global Optimization} 
& \multicolumn{6}{c}{with Global Optimization} \\
\cmidrule(lr){2-7} \cmidrule(lr){8-13}
& \multicolumn{3}{c}{Single-dance} & \multicolumn{3}{c|}{Group-dance} 
& \multicolumn{3}{c}{Single-dance} & \multicolumn{3}{c}{Group-dance} \\
\cmidrule(lr){2-4}\cmidrule(lr){5-7}\cmidrule(lr){8-10}\cmidrule(lr){11-13}
& FID$\downarrow$ & Div$\uparrow$ & PFC$\downarrow$
& GMR$\downarrow$ & GMC$\uparrow$ & TIF$\downarrow$
& FID$\downarrow$ & Div$\uparrow$ & PFC$\downarrow$
& GMR$\downarrow$ & GMC$\uparrow$ & TIF$\downarrow$ \\
\midrule
EDGE (AIST++)      & \underline{24.74} & 12.98 & 1.04 & 15.23 & 80.21 & 0.13 
                   & \underline{24.74} & 12.98 & 1.04 & 12.32 & 82.49 & 0.11 \\
EDGE (FineDance)   & 26.12 & 13.34 & \underline{0.62} & 14.93 & 80.73 & \underline{0.12} 
                   & 26.12 & 13.34 & \underline{0.62} & 12.25 & 82.65 & \textbf{0.10} \\
Lodge (AIST++)     & 27.31 & \underline{14.41} & 1.01 & \underline{14.63} & \underline{80.99} & \underline{0.12} 
                   & 27.31 & \underline{14.41} & 1.01 & \underline{11.69} & \underline{83.30} & 0.11 \\
Lodge (FineDance)  & \textbf{23.12} & \textbf{13.85} & \textbf{0.50} & \textbf{12.44} & \textbf{82.32} & \textbf{0.11} 
                   & \textbf{23.12} & \textbf{13.85} & \textbf{0.50} & \textbf{11.63} & \textbf{83.74} & \textbf{0.10} \\
\bottomrule
\end{tabular}}
\end{table}
\vspace{-12pt}
Table~\ref{tab:backbone_dataset} summarizes results across different backbones and pretraining datasets. The global noise optimization defines penalty rules over inter-dancer interactions and therefore does not affect single-dance generation. Overall, adding this global penalty improves group dance (3 dancers) quality in most settings, particularly for GMR and GMC. For example, with the Lodge backbone on FineDance, GMR decreases from 12.44 to 11.63 and GMC increases from 82.32 to 83.74, showing better spatial coordination and rhythm alignment.
The EDGE backbone is more efficient but produces slightly lower quality than Lodge. Models trained on FineDance outperform those trained on AIST++, likely due to FineDance's broader style coverage and richer motion variety. These results show that ChainDance successfully adapts single-dancer models for stable multi-person generation across different architectures and datasets.

\vspace{-10pt}
\section{Conclusion}
\vspace{-6pt}
We presented \textbf{ChainDance}, a parameter-efficient framework that scales group dance generation by reformulating the task as a \textit{Chain-of-Dancers}. By adapting a frozen backbone with two lightweight modules, a \textbf{Role-Aware Text Encoder} and a \textbf{Group-Aware Motion Encoder}, our model synthesizes variable group sizes through pair-wise conditioning and inference-time noise optimization. On AIOZ-GDance, ChainDance achieves state-of-the-art motion quality while structurally preserving each dancer's identity, using $3$-$4\times$ fewer parameters and $3$-$6\times$ less training time. More broadly, our results suggest that structured decomposition is a viable alternative to joint modeling for multi-agent motion generation.

\begin{ack}
\textbf{Funding.} This research was supported by the Australian Government through the Australian Research Council Discovery Early Career Researcher Award (DE250100030) and Discovery Project (DP260100218). Qiuhong Ke is the recipient of the Discovery Early Career Researcher Award (DE250100030).

\textbf{Competing interests.} The authors declare no competing interests.
\end{ack}

\bibliographystyle{plainnat}
\bibliography{ref} 


\appendix
\section{Prior Assumptions: Joint Distribution Perspective on Group Dance Generation}
\label{sec:assumptions}
From a probabilistic modeling perspective, generative models fundamentally learn conditional probability distributions over data. 
For the group dance generation task considered in this work, let $\mathcal{X}_{1:N}=\{X_1, X_2, \dots, X_N\}$ denote a set of dance motion sequences performed by $N$ dancers, where each $X_i$ represents the full temporal motion trajectory of the $i$-th dancer. 
The learning objective of group dance generation can thus be formulated as modeling the joint distribution $P(\mathcal{X}_{1:N})$.

According to the chain rule of probability, the joint distribution can be exactly factorized as:
\begin{equation}
P(\mathcal{X}_{1:N}) = P(X_1)\prod_{i=2}^{N} P(X_i \mid \mathcal{X}_{1:i-1}),
\end{equation}
which is a mathematically equivalent decomposition and does not introduce any approximation.

Our approach is directly motivated by this factorization. 
We first leverage a well-trained and stable single-dancer diffusion model to capture the marginal distribution $P(X_1)$, providing strong temporal dynamics and motion priors. 
For each subsequent dancer $X_i$, we model the conditional distribution $P(X_i \mid \mathcal{X}_{1:i-1})$ via conditional generation, where the conditioning signals consist of the motions of previously generated dancers together with multimodal control inputs.

This formulation offers two key advantages. 
First, temporal continuity, rhythm, and motion realism---already well captured by the single-dance generator---do not need to be relearned in the group setting. 
Second, inter-dancer interactions and spatial constraints are explicitly introduced through conditional dependencies during generation, avoiding the need to directly model the full high-dimensional joint distribution in a single step.

Therefore, from the perspective of joint probability modeling, the proposed sequential conditional generation strategy preserves the expressiveness of the target group dance distribution while providing a structured and scalable approach that improves training stability and generation robustness.

\section{Implementation Details}
\label{sec:loss_details}
This section provides the detailed formulations of graph construction, all loss terms and encoder details related to ChainDance.
Each loss term corresponds to either the local reconstruction stage or the global refinement stage described in Section~\ref{sec:method} of the main paper.

\subsection{Group-Aware Motion Graph}
\label{sec:graph_construction}

The Group-Aware Motion Encoder (Section~\ref{sec:method}) is implemented as a graph convolutional network operating on a \emph{distance-weighted dynamic graph} that grows incrementally as new dancers are generated.

\paragraph{Distance-weighted edges.}
For the $i$-th dancer (node) being generated, we define raw interaction weights $w_{ij}$ with all previously generated dancers $j$ based on their Euclidean distance:
\begin{equation}
w_{ij} = \frac{1}{|p_i - p_j| + \epsilon},
\label{eq:graph_weight}
\end{equation}
where $p_i$ and $p_j$ denote the global root positions of dancer $i$ and $j$, respectively, and $\epsilon = 10^{-6}$ is a small constant added to avoid division by zero. This formulation encodes a natural inductive bias: \emph{closer dancers exert a stronger influence on the generation of the current dancer}, mirroring how real choreography prioritizes nearby partners over distant ones.

\paragraph{Graph construction.}
Following the chain-of-dancers formulation, the graph is built incrementally rather than over a fully connected static topology. When a new dancer is added to the chain, edges are dynamically constructed only between the new node and the previously generated dancers, while existing edges are preserved. This design has been shown effective in recent graph-based generation studies, including I-GCN~\cite{nie2021gcn} and IGC~\cite{ding2022causal}, and naturally fits our sequential generation scheme. During diffusion sampling, the follower's root position is taken from the intermediate prediction $\hat{X}_i^{(t)}$ at each timestep $t$; the graph is recomputed dynamically as denoising progresses.

\paragraph{Top-$k$ edge sparsification.}
To focus on local coordination and improve computational efficiency, we sparsify the graph by retaining only the top $60\%$ nearest edges (i.e., highest $w_{ij}$ values) for each newly added dancer. This avoids over-smoothing from far-away dancers whose contributions to local coordination are minimal, while preserving sufficient connectivity to model meaningful inter-dancer relationships. The sparsified weights are then row-normalized via softmax to obtain the final adjacency matrix used in the GCN layer:
\begin{equation}
\tilde{w}_{ij} = \frac{\exp(w_{ij})}{\sum_{k \in \mathcal{N}_i} \exp(w_{ik})},
\end{equation}
where $\mathcal{N}_i$ denotes the set of retained neighbors of dancer $i$ after top-$k$ sparsification.

\subsection{Algorithm Details}
\label{sec:alg_details}
Due to space limitations in the main paper, we provide a detailed description of the Group-Aware Motion Encoder in Algorithm~\ref{alg:inter_dancer_adapter}.
This section explicitly specifies the input and output representations, intermediate feature dimensionalities, and the sequence of modules used to generate inter-dancer motion-aware key/value pairs.
The goal is to clarify how motion interactions are encoded and injected into the frozen single-dancer diffusion backbone, complementing the high-level description in the main text.

\begin{algorithm}[H]
\caption{Group-Aware Motion Encoder}
\label{alg:inter_dancer_adapter}
\small
\begin{algorithmic}
\STATE \textbf{Input:} Immediate leader motion $\mathbf{X}_j$ (where $j = i-1$), all previously generated motions $\{\mathbf{X}_1, \ldots, \mathbf{X}_{i-1}\}$
\STATE \textbf{Output:} Inter-dancer key/value $(\mathbf{K}_{grp}, \mathbf{V}_{grp})$
\STATE \quad $\mathcal{G} \leftarrow \text{BuildGraph}(\{\mathbf{X}_1, \ldots, \mathbf{X}_{i-1}\})$
\hfill // Distance-weighted graph over all previously generated dancers
\STATE \quad $\mathbf{Z}_j \leftarrow f_{\text{motion}}(\mathbf{X}_j)$
\hfill // Leader Motion Encoding, {$\mathbf{X}_j \in \mathbb{R}^{B \times T \times D_m},
          \mathbf{Z}_j \in \mathbb{R}^{B \times T \times d_z}$}
\STATE \quad $\tilde{\mathbf{Z}}_j \leftarrow \text{GCN}(\mathbf{Z}_j, \mathcal{G})$
\hfill // Aggregate group-wide spatial context onto leader feature, temporal length $L$ preserved
\STATE \quad $\mathbf{H}_j \leftarrow \text{Transformer}_{\text{LoRA}}(\tilde{\mathbf{Z}}_j)$
\hfill // LoRA-Adapted Transformer Encoding, which applies low-rank updates to linear projections
\STATE \quad $\mathbf{K}_{grp} \leftarrow \mathbf{H}_j \mathbf{W}_K$
\hfill // Key / Value Projection, $\mathbf{K}_{grp}, \mathbf{V}_{grp} \in \mathbb{R}^{B \times H \times L \times d_h}$
\STATE \quad $\mathbf{V}_{grp} \leftarrow \mathbf{H}_j \mathbf{W}_V$
\STATE \quad Inject $(\mathbf{K}_{grp}, \mathbf{V}_{grp})$ into frozen backbone
\hfill // Cross-Attention Injection
\end{algorithmic}
\end{algorithm}

\subsection{Training Loss}
The detailed definitions of the loss terms used in Equation~\ref{eq:local} in Section~\ref{sec:chaingen} are given as follows.

\paragraph{Pose Reconstruction Loss.}~To ensure accurate recovery of dancer poses, we use an $\ell_1$ loss between predicted $\hat{\mathbf{X}}^{(n)}_l$ and ground-truth $\mathbf{X}^{(n)}_l$ joint representations:
\begin{equation}
\mathcal{L}_{\mathrm{pose}}
= \frac{1}{NL}\sum_{n=1}^{N}\sum_{l=1}^{L}
\left\| \hat{\mathbf{X}}^{(n)}_l - \mathbf{X}^{(n)}_l \right\|_1.
\end{equation}
This term drives the model to reproduce spatially precise skeleton structures at each frame.

\paragraph{Velocity Regularization Loss.}~To suppress temporal jitter and enforce smooth motion transitions, we apply a velocity consistency constraint:
\begin{equation}
\mathcal{L}_{\mathrm{vel}}
= \frac{1}{NL}\sum_{n,l}
\left\| (\hat{\mathbf{X}}^{(n)}_{l} - \hat{\mathbf{X}}^{(n)}_{l-1})
      - (\mathbf{X}^{(n)}_{l} - \mathbf{X}^{(n)}_{l-1}) \right\|_1.
\end{equation}
It encourages the predicted motion derivatives to follow the ground-truth temporal dynamics.

\paragraph{Foot-Ground Contact Loss.}~To reduce foot sliding and floating artifacts, we constrain the contact joints (e.g., ankles and toes) relative to the ground plane:
\begin{equation}
\mathcal{L}_{\mathrm{contact}}
= \frac{1}{L}\sum_{l}
\sum_{k\in\Omega_{\mathrm{foot}}}
\left(
\left| \hat{z}_{l,k} \right| +
\alpha \left| \hat{v}_{l,k} \right|
\right),
\end{equation}
where $\hat{z}_{l,k}$ and $\hat{v}_{l,k}$ denote the height and velocity of foot joint $k$, and $\alpha$ balances positional and velocity constraints.

\paragraph{Multimodal Alignment Loss.}~We follow the text-pivoted alignment formulation proposed in DanceChat~\cite{wang2025dancechat}, which encourages consistency between modalities by aligning both text--music and text--motion embeddings via cosine similarity.
At each timestep $i$, the alignment loss is defined as:
\begin{equation} \mathcal{L}_{\mathrm{align}} =\frac{1}{N}\sum_{i=1}^{N} \Big[(1-\mathrm{sim}(E_{t_i},E_{m_i})) +(1-\mathrm{sim}(E_{t_i},E_{x_i}))\Big], 
\end{equation} 
where $\mathrm{sim}(\mathbf{a},\mathbf{b})=\tfrac{\langle\mathbf{a},\mathbf{b}\rangle}{\|\mathbf{a}\|\,\|\mathbf{b}\|}$ denotes cosine similarity and $E_{t_i}$, $E_{m_i}$, $E_{x_i}$ represent the encoded embeddings of text, music, and motion, respectively, at timestep $i$.
While we adopt the same loss structure as DanceChat, we re-implement the encoders $E_m$ and $E_x$ using our own music and motion backbones to ensure compatibility with the encoder-based architecture.

\subsection{Inference Global Penalties}
We used GPU parallel acceleration in different sample batch dimensions to minimize latency, where multiple noise samples are optimized in parallel within each iteration of gradient descent. In Section~\ref{sec:globalopt}, we introduce a global penalty at inference time. The loss terms of $L_{\mathrm{global}}$ in Equation~\ref{eq:global_controls} are listed as follows:

\paragraph{Root Position Penalty.}
To ensure that an individual reaches a specific location at a particular time, 
we define the root position penalty $L_{\text{root}}$ as follows:
\begin{equation}
L_{\text{root}}=\sum_{l\in L} \max\left(0, \|p_{\text{root}}^{p}(l)-p_{\text{target}}(l)\|^{2}-\delta\right),
\label{eq:root-penalty}
\end{equation}
where $p_{\text{root}}^{p}(l)$ represents the root position of individual $p$ at frame $l$, 
and $p_{\text{target}}(l)$ is the target position at the same frame. 
The hyperparameter $\delta$ acts as a threshold distance, below which no penalty is applied. This formulation relaxes the loss function by penalizing deviations from the target position 
only when the distance exceeds $\delta$. Such a design avoids overly rigid constraints on 
close positions and encourages more natural and flexible motion synthesis, consistent with 
prior work PINO~\cite{ota2025pino}.

\begin{table*}[t]
\centering
\small
\setlength{\tabcolsep}{6pt}
\begin{tabularx}{\textwidth}{@{}l X l@{}}
\toprule
\textbf{Relation (REL)} & \textbf{Decision rule in window $\mathcal{W}$} & \textbf{Artifacts (for prompt)} \\
\midrule
\texttt{approach} &
Monotonic decrease in distance: $\Delta d/W < -0.1$ and
$d_{ij}(w_0+W-1)\in[d_{\text{near}},\,d_{\text{mid}}]$. &
WHEN, target distance $d^\star\!\approx d_{\text{near}}$ \\
\addlinespace[2pt]
\texttt{keep\_distance} &
Low variance around mid distance:
$\mathrm{Var}_{w\in\mathcal{W}}(d_{ij}(w))<0.05$ and
$\mathrm{mean}(d_{ij})\in[d_{\text{mid}}-0.2,\,d_{\text{mid}}+0.2]$. &
WHEN, target distance $d^\star\!\approx d_{\text{mid}}$ \\
\addlinespace[2pt]
\texttt{face\_to} &
Mutual facing with proximity:
$\cos\!\big(\mathbf{f}_i,\mathbf{p}_j-\mathbf{p}_i\big)>0.7$ and
$\cos\!\big(\mathbf{f}_j,\mathbf{p}_i-\mathbf{p}_j\big)>0.7$ and
$d_{ij}<d_{\text{far}}$. &
WHEN \\
\addlinespace[2pt]
\texttt{lead\_follow} &
Cross-correlation of motion energies peaks at a positive lag:
$\arg\max_{\Delta w\in[2,8]}\mathrm{xcorr}(E_i,E_j)=\Delta w>0$. (Here $i$ leads, $j$ follows.) &
WHEN, lag $\Delta w$; leader/follower IDs \\
\addlinespace[2pt]
\texttt{swap} &
Start/end positions swap across the window:
$\|\mathbf{p}_i(w_0)-\mathbf{p}_j(w_0+W-1)\|_2<\epsilon$ and
$\|\mathbf{p}_j(w_0)-\mathbf{p}_i(w_0+W-1)\|_2<\epsilon$,
while trajectories do not collide. &
WHEN \\
\addlinespace[2pt]
\texttt{passby} &
Trajectories intersect near mid/near distance with opposite normal velocities:
$\exists\,w\in\mathcal{W}$ s.t.\ $d_{ij}(w)\in[d_{\text{near}},\,d_{\text{mid}}]$,
$\langle\dot{\mathbf{p}}_i^\perp,\dot{\mathbf{p}}_j^\perp\rangle<0$,
and distance increases rapidly after intersection. &
WHEN \\
\bottomrule
\end{tabularx}
\caption{Window-based relation detection rules for prompt mining.
\emph{WHEN} records the time span(s) in $\mathcal{W}$ where the rule is satisfied;
artifacts (e.g., $d^\star$, contacted parts, lag $\Delta t$) are injected into prompt templates.}
\label{tab:rel_rules}
\end{table*}

\paragraph{Movement Region Penalty.}
To restrict movement within a defined area, we introduce the movement region penalty 
$L_{\text{region}}$ for a chosen individual $p$:
\begin{equation}
L_{\text{region}} = \frac{1}{LJ} \sum_{n} \sum_{j} \,\phi\left(p^{p}_{j}(n)\right),
\label{eq:region-penalty}
\end{equation}
where $L$ is the total number of frames, $J$ is the total number of joints, 
and $\phi(\cdot)$ penalizes positions outside the desired region.  
Here, $p^{p}_{j}(n)$ denotes the position of joint $j$ of individual $p$ at frame $n$. The penalty function $\phi(\cdot)$ can be flexibly defined according to task requirements. 
For example, restricting movement within a rectangular cuboid region with lower bounds 
$l=(l_x,l_y,l_z)$ and upper bounds $u=(u_x,u_y,u_z)$ yields:
\begin{equation}
\phi(p) = \sum_{k\in\{x,y,z\}} \max(0,\, l_k - p_k) + \max(0,\, p_k - u_k),
\label{eq:phi-cuboid}
\end{equation}
where $p_k$ is the $k$-th coordinate of $p$.

\vspace{-4pt}
\paragraph{Orientation Penalty.}
To control facing directions at specific frames, we define the orientation penalty 
$L_{\text{orient}}$ with threshold $\delta$ as:
\begin{equation}
L_{\text{orient}}=
\sum_{l\in L}\max\left(0,\,1 - d^{p}(l)\cdot d_{\text{target}}(l) - \delta\right),
\label{eq:orient-penalty}
\end{equation}
where $d^{p}(l)$ is the normalized facing direction of individual $p$ at frame $l$, 
and $d_{\text{target}}(l)$ is the desired direction.  
The threshold $\delta$ determines the tolerance margin within which no penalty is applied. By penalizing deviations only when direction similarity falls below $1-\delta$, this 
formulation promotes natural orientation control and avoids abrupt directional changes, 
similar to the soft-direction constraints adopted in PINO~\cite{ota2025pino}.

\paragraph{Relative Position Penalty.}
To ensure that the root positions of two individuals remain within a desired distance range, 
we define the relative position penalty $L_{\text{relative}}$ as:
\begin{equation}
\begin{aligned}
L_{\text{relative}}
= \sum_{l\in L} &\Big[
\max\left(0,\, d_{\min}-\|p_{\text{root}}^{1}(l) - p_{\text{root}}^{2}(l)\|\right) \\
& + \max\left(0,\, \|p_{\text{root}}^{1}(l) - p_{\text{root}}^{2}(l)\| - d_{\max}\right)
\Big],
\end{aligned}
\label{eq:relative-penalty}
\end{equation}
where $p_{\text{root}}^{1}(l)$ and $p_{\text{root}}^{2}(l)$ denote the root positions 
of two individuals at frame $l$, and $d_{\min}$ and $d_{\max}$ represent the lower and 
upper bounds of the allowable interpersonal distance. This constraint maintains natural interaction spacing and prevents unrealistic crowding or separation, an important requirement also emphasized by PINO~\cite{ota2025pino}. 

To accelerate the optimization of global penalty over initial noise at the inference stage, in the main experiment, we did not use the penalty of the entire long-time sequence as the optimization target, but instead selected the 70 frames with the largest penalty values in the entire sequence as the optimization target. Meanwhile, we used GPU parallel acceleration in different sample batch dimensions to minimize latency in the inference stage.

\section{From Rules to Interaction Types}
\label{sec:rule_interaction}

As described in Section~\ref{sec:textprompt} and illustrated in Figure~\ref{fig:textprompt}, 
the mined pairwise relations from each window $\mathcal{W}$ are verbalized into 
dancer-specific textual instructions that serve as the conditioning input for ChainDance. 
We use a sliding temporal window $\mathcal{W}$ to define these relations: for each window, 
we look at the locations of two dancers within this short time segment and decide whether an interaction occurs and, if so, what type it is. This design uses windows instead of the whole sequence because 
interactions do not happen at every frame and are usually completed within a short time range. 
We adopt minimal, compositional templates that consume the artifacts produced by the 
decision rules (e.g., estimated distance $d^\star$, lag $\Delta t$, and time span \emph{WHEN}), 
while strictly following the prompt format constraints in Section~\ref{sec:textprompt}. The detailed computation of these rules is summarized in Table~\ref{tab:rel_rules}.

\vspace{-6pt}
\paragraph{Templates.}~Detected relations are mapped to short phrases as follows:
\begin{itemize}[leftmargin=1.2em,itemsep=0pt,topsep=2pt]
  \item \texttt{approach}: ``\emph{Dancer-$i$ approaches Dancer-$j$ to about $d^\star$.}''
  \item \texttt{keep\_distance}: ``\emph{$i$ and $j$ keep a steady distance $\approx d^\star$.}''
  \item \texttt{face\_to}: ``\emph{$i$ and $j$ face each other.}''
  \item \texttt{lead\_follow}: ``\emph{$i$ leads $j$ (lag $\Delta w$).}''
  \item \texttt{swap}: ``\emph{$i$ and $j$ swap positions at.}''
  \item \texttt{passby}: ``\emph{$i$ and $j$ pass by.}''
\end{itemize}
These phrases are then injected into per-dancer prompts 
(e.g., ``\emph{The dancer $i$ \ldots, facing toward Dancer $j$.}'') on top of the global 
description, yielding individualized conditioning as specified in Section~\ref{sec:textprompt}.

\vspace{-10pt}
\paragraph{De-duplication and Priority.}~If multiple relations fire in the same window, we keep the one with the highest confidence 
(e.g., larger margin to threshold) and merge contiguous \emph{WHEN} intervals if gaps are $<\!0.5$\,s.
The final per-dancer prompt is constructed by (i) preserving the global motif and (ii) appending at most 
one relation phrase per active pair within $\mathcal{W}$. 

\vspace{-8pt}
\section{More Quantitative Results}
\label{additional_quantitative}
\subsection{Comparison of 5 Dancers}

\begin{table*}[t]
\centering
\caption{
\textbf{Unified comparison across 1–5 dancers on the AIOZ\mbox{-}GDance dataset.}
For all baselines, \emph{separate models} are trained and evaluated for each group size.
In contrast, \textbf{ChainDance uses a single unified model for all group sizes (1–5 dancers)}.
Best results are \textbf{bold}; second-best are \underline{underlined}. 
Results with * denote models without the global noise optimization penalty.
}
\label{tab:all_dancer_unified}
\setlength{\tabcolsep}{4.6pt}
\renewcommand{\arraystretch}{1.12}
\resizebox{\textwidth}{!}{
\begin{tabular}{l|ccc|ccc|ccc|ccc|ccc}
\hline
\multirow{2}{*}{Method} &
\multicolumn{3}{c|}{\textbf{1 dancer}} &
\multicolumn{3}{c|}{\textbf{2 dancers}} &
\multicolumn{3}{c|}{\textbf{3 dancers}} &
\multicolumn{3}{c|}{\textbf{4 dancers}} &
\multicolumn{3}{c}{\textbf{5 dancers}} \\
\cline{2-16}
& FID$\downarrow$ & Div$\uparrow$ & PFC$\downarrow$
& GMR$\downarrow$ & GMC$\uparrow$ & TIF$\downarrow$
& GMR$\downarrow$ & GMC$\uparrow$ & TIF$\downarrow$
& GMR$\downarrow$ & GMC$\uparrow$ & TIF$\downarrow$
& GMR$\downarrow$ & GMC$\uparrow$ & TIF$\downarrow$ \\
\hline
GCD~\cite{le2023controllable}
& 39.24 & 9.64 & 2.53
& 34.39 & 80.32 & 0.17
& 30.22 & 80.22 & 0.19
& 36.28 & 81.82 & 0.13
& 38.43 & 81.44 & 0.17 \\

CoDancers~\cite{yang2024codancers}
& \underline{23.98} & 9.48 & 3.53
& 24.55 & 72.52 & \textbf{0.08}
& 26.34 & 74.22 & \textbf{0.08}
& 26.44 & 75.34 & \textbf{0.10}
& 27.27 & 74.34 & \textbf{0.11} \\

TCDiff~\cite{dai2025harmonious}
& 37.31 & \textbf{14.01} & \underline{0.51}
& 15.77 & \textbf{81.92} & 0.12
& 15.36 & \underline{82.77} & 0.11
& 13.44 & 81.70 & 0.15
& \textbf{14.62} & 81.40 & \textbf{0.11} \\

ST\mbox{-}GDance~\cite{xu2025stgdance}
& 28.87 & 12.82 & 0.97
& 19.42 & 80.52 & 0.12
& 14.76 & 81.24 & 0.11
& 14.02 & 81.42 & \textbf{0.10}
& 23.22 & 80.76 & \underline{0.12} \\
\hline
ChainDance* 
& \textbf{23.12} & \underline{13.85} & \textbf{0.50}
& \underline{14.10} & \underline{81.52} & \underline{0.11}
& \underline{12.01} & 82.68 & 0.11
& \underline{13.28} & \underline{82.13} & \underline{0.12}
& 15.34 & \underline{80.93} & 0.13 \\

ChainDance
& \textbf{23.12} & \underline{13.85} & \textbf{0.50}
& \textbf{13.67} & 81.28 & 0.12
& \textbf{11.63} & \textbf{83.74} & \underline{0.10}
& \textbf{12.59} & \textbf{82.16} & 0.13
& \underline{14.73} & \textbf{81.56} & \underline{0.12} \\
\hline
\end{tabular}
}
\end{table*}

Table~\ref{tab:all_dancer_unified} summarizes the results across the 1--5 dancer settings. We report 5-dancer results here rather than in the main paper because the number of 5-dancer samples in the AIOZ-GDance test set is limited, making the comparison less statistically robust. At $N{=}5$, ChainDance achieves the best GMC (\textbf{81.56}) and the second-best GMR (14.73 vs.\ TCDiff's 14.62), with the small GMR gap reflecting a fundamental trade-off of our design: while joint-modeling baselines can be specifically optimized for each fixed group size, ChainDance's unified pair-wise training transfers across all sizes from a single model. Across the full 1--5 dancer range, ChainDance remains consistently competitive with one unified model, an advantage no baseline can match.

\subsection{Comparison on DD100 dataset}
\label{sec:dd100}

Although AIOZ-GDance is the only publicly available large-scale group dance benchmark, several duet datasets exist. We therefore evaluate ChainDance's generalization on DD100~\cite{siyaoduolando}, a duet dance dataset with 117 minutes of professional motion-capture performances across 10 genres. Following our main setup, we adopt the Lodge~\cite{li2024lodge} backbone pretrained on FineDance~\cite{li2023finedance} and post-train the group-aware modules on DD100. Since DD100 does not provide text annotations, we disable the Role-Aware Text Encoder for this experiment and condition only on music, while retaining the Group-Aware Motion Encoder and chain-of-dancers formulation. We follow the official train/test split and compare with Duolando~\cite{siyaoduolando} and GCD~\cite{le2023controllable} under the same evaluation protocol. Results are reported in Table~\ref{tab:dd100}.

\noindent As shown in Table~\ref{tab:dd100}, ChainDance achieves the best performance across all metrics on DD100, demonstrating that our framework generalizes effectively beyond AIOZ-GDance. In particular, ChainDance reduces FID from 9.73 (GCD) to \textbf{2.97}, achieves the closest Div to GT (15.52 vs.\ GT's 15.67), and outperforms both Duolando and GCD on group-level metrics. These results confirm that even with text conditioning disabled, the chain-of-dancers formulation, combined with the Group-Aware Motion Encoder, generalizes effectively to the duet setting.

\begin{table}[t]
\centering
\caption{\textbf{Two-dancer generation results on the DD100 dataset.} Best results are \textbf{bold}. ``$\rightarrow$'' indicates that closer to ground-truth (GT) is better.}
\label{tab:dd100}
\footnotesize
\setlength{\tabcolsep}{6pt}
\renewcommand{\arraystretch}{1.15}
\begin{tabular}{lcccccc}
\toprule
Method & FID$\downarrow$ & Div$\rightarrow$ & PFC$\downarrow$ & GMR$\downarrow$ & GMC$\uparrow$ & TIF$\downarrow$ \\
\midrule
GT          & --              & 15.67              & 0.36              & --                  & --                  & -- \\
\midrule
Duolando~\cite{siyaoduolando} & 12.42 & 14.35 & 16.22 & 14.52 & 81.32 & 0.15 \\
GCD~\cite{le2023controllable} & 9.73 & 14.62 & 5.12 & 14.31 & 81.21 & 0.13 \\
\textbf{ChainDance (w/o RATE)}           & \textbf{2.97} & \textbf{15.52} & \textbf{4.41} & \textbf{12.21} & \textbf{82.42} & \textbf{0.12} \\
\bottomrule
\end{tabular}
\end{table}

\subsection{Comparison with PINO}
A potential concern is whether the gains of ChainDance over end-to-end baselines simply come from the test-time noise optimization adopted from PINO~\cite{ota2025pino}. To address this, we directly compare ChainDance with PINO using the same single-dancer backbone for group sizes 2--4, isolating the contribution of our proposed encoders and chain-of-dancers formulation.

\begin{table}[b]
\centering
\caption{\textbf{Comparison with PINO across different group sizes.} Best results are in \textbf{bold}.}
\label{tab:pino_compare}
\footnotesize
\setlength{\tabcolsep}{4.5pt}
\renewcommand{\arraystretch}{1.1}
\begin{tabular}{lccc|ccc|ccc}
\toprule
\multirow{2}{*}{Method} &
\multicolumn{3}{c|}{\textbf{2}} &
\multicolumn{3}{c|}{\textbf{3}} &
\multicolumn{3}{c}{\textbf{4}} \\
\cmidrule(lr){2-4}\cmidrule(lr){5-7}\cmidrule(lr){8-10}
& GMR$\downarrow$ & GMC$\uparrow$ & TIF$\downarrow$
& GMR$\downarrow$ & GMC$\uparrow$ & TIF$\downarrow$
& GMR$\downarrow$ & GMC$\uparrow$ & TIF$\downarrow$ \\
\midrule
PINO~\cite{ota2025pino}
& 14.10 & 80.83 & 0.12
& 12.69 & 82.92 & 0.11
& 13.83 & 80.93 & 0.14 \\
\textbf{ChainDance}
& \textbf{13.67} & \textbf{81.28} & \textbf{0.12}
& \textbf{11.63} & \textbf{83.74} & \textbf{0.10}
& \textbf{12.59} & \textbf{82.16} & \textbf{0.13} \\
\bottomrule
\end{tabular}
\end{table}

\noindent As shown in Table~\ref{tab:pino_compare}, ChainDance consistently outperforms PINO across all group sizes. Notably, the performance gap \emph{widens} as the group size increases (e.g., GMR improves by 0.43 for 2 dancers but by 1.24 for 4 dancers), indicating that relying solely on global physical constraints becomes increasingly insufficient as inter-dancer interactions grow more complex. While PINO refines spatial consistency at inference time, it lacks the capacity to model fine-grained inter-dancer relationships during generation. Our encoders address this gap by injecting relational information directly into the latent space, providing the structural foundation that PINO's optimization alone cannot establish. This isolation experiment confirms that ChainDance's contribution goes beyond inference-time refinement: the proposed encoders and chain-of-dancers formulation are the key drivers of the observed performance gains.

\subsection{User Study}
\label{sec:user_study}

Beyond automatic metrics, we conduct a user study to evaluate the perceptual quality of generated group dances. Following the protocol of our main baseline TCDiff~\cite{dai2025harmonious}, we invite \textbf{30 participants} (ages 18--55, aligning with the target user group of our application scenario) to participate in a blind ranking test. Outputs from all methods are anonymized and shuffled before presentation. For each comparison, participants are shown four videos (one per method) and asked to rank them from best to worst along four perceptual dimensions:

\begin{itemize}[leftmargin=*, itemsep=2pt, topsep=2pt]
    \item \textbf{Realism}: How natural and human-like are the generated motions?
    \item \textbf{Music Correlation}: How well do the dance movements align with the music's rhythm and style?
    \item \textbf{Formation Aesthetics}: How visually pleasing and well-coordinated are the group formations?
    \item \textbf{Dancer Harmony}: How harmoniously do the dancers interact and complement each other, and how consistently does each dancer maintain a stable identity throughout the sequence (i.e., absence of identity swaps or sudden position jumps)?
\end{itemize}

\begin{figure}[t]
  \centering
  \includegraphics[width=0.8\linewidth]{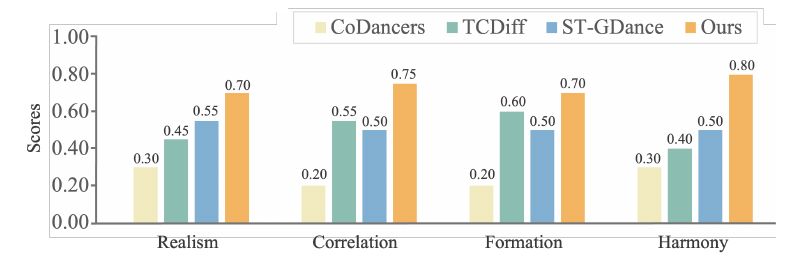}
  \caption{\textbf{User study results across four perceptual dimensions.} ChainDance is consistently preferred over baselines on Realism, Music Correlation, Formation Aesthetics, and Dancer Harmony. The largest improvement is observed on Dancer Harmony (0.80 vs.\ 0.50 of the best baseline), which jointly captures inter-dancer coordination and per-dancer identity stability---both structurally improved by our sequential decomposition.}
  \label{fig:userstudy}
\end{figure}

Rankings are converted into scores from 4 (best) to 1 (worst), and the averaged scores are further normalized to $[0,1]$ for fair comparison. Results are reported in Figure~\ref{fig:userstudy}.

\noindent As shown in Figure~\ref{fig:userstudy}, ChainDance is preferred across all four dimensions, achieving the highest normalized scores on Realism (0.70), Music Correlation (0.75), Formation Aesthetics (0.70), and Dancer Harmony (0.80). The largest gap appears on \emph{Dancer Harmony}, where ChainDance scores 0.80 compared to 0.50 for the strongest baseline on this dimension (ST-GDance). This dimension captures both inter-dancer coordination and per-dancer identity stability over time---two aspects that joint-modeling baselines struggle with simultaneously, as joint tokenization tends to produce identity swaps and flickering artifacts (Figure~\ref{fig:challenges}b). ChainDance's sequential decomposition structurally avoids identity entanglement, while RATE and GAME ensure coherent inter-dancer interaction. This provides direct perceptual evidence supporting the identity preservation claim, complementing the objective metrics in Table~\ref{tab:groupdance} and Table~\ref{tab:pino_compare}, jointly demonstrating that ChainDance produces group choreography that is both quantitatively superior \emph{and} perceptually preferred.

\subsection{Other Ablation Settings}
\label{otherabla}
To comprehensively evaluate ChainDance, we compare results under different configurations: single-dancer generators pretrained on AIOZ-GDance, models pretrained jointly on FineDance and AIST++, and variants trained without motion captions in text descriptions.

Table~\ref{tab:otherdataset} summarizes several additional ablation settings. When we pretrain the single-dancer generator on the AIOZ-GDance subset and then finetune with multi-dancer clips, the performance is slightly lower than the main results in the paper. This is understandable because AIOZ-GDance is primarily a multi-person dataset, and its single-dance samples are limited, which makes the single-dancer pretraining less effective.

\begin{table}[H]
\centering
\caption{\textbf{Comparison of 3 dancers performance under different ablation settings on the AIOZ\mbox{-}GDance dataset.} Best results are \textbf{bold}.}
\label{tab:otherdataset}
\setlength{\tabcolsep}{4.5pt}
\renewcommand{\arraystretch}{1.12}
\small
\begin{tabular}{l|ccc}
\hline
\multirow{2}{*}{Method} &
\multicolumn{3}{c}{\textbf{Group Dance}} \\
\cline{2-4}
& GMR$\downarrow$ & GMC$\uparrow$ & TIF$\downarrow$ \\
\hline
ChainDance
& \textbf{11.63} & \textbf{83.74} & \textbf{0.10} \\
\hline
Ours-Lodge(AIOZ-GDance)
& 12.08 & 82.32 & 0.11\\
Ours-Lodge (FineDance\&AIST++)
& 11.97 & 82.41 & 0.12 \\
\hline
ChainDance (w/o Motion Caption)
& 11.89 & 82.52 & 0.11 \\

ChainDance (w/o bidirectional pairing)
& 12.93 & 79.42 & 0.13 \\

ChainDance (with random pairing strategy)
& 11.58 & 83.80 & 0.10 \\
\hline
\end{tabular}
\end{table}

Using a backbone pretrained jointly on FineDance and AIST++ achieves similar performance to the AIOZ-GDance-based model, but remains below the version pretrained only on FineDance in the main paper. A possible explanation is that the mixed-domain training increases the diversity of styles and camera motions, which benefits generality but introduces additional variance that is not fully aligned with AIOZ-GDance.

Removing motion captions from the text descriptions leads to a small drop in performance, but the results remain stable and still compare favorably with end-to-end multi-dancer baselines that do not use such information. This suggests that motion captions are helpful, but they are not the dominant factor. The improvements mainly come from the full text inputs (music, motion caption, beat structure) together with the ChainDance formulation, which simultaneously provide the model with strong guidance and allow efficient multi-dancer generation. Training pairs without bidirectional leader relationship between dancers may cause the model to learn only some fixed local interaction patterns, thus greatly reducing the effectiveness of group dance generation. Comparing the two pairing strategies, both random pairing and chain pairing yield similar group-level performance, with only marginal differences across metrics. For clarity and reproducibility, we use chain pairing as the default setting in the main experiments. 

\subsection{Error Accumulation Analysis Across Group Sizes}
\label{sec:error_accum}

A natural concern with sequential chain generation is whether errors accumulate 
as group size grows. Our inference-time global optimization is designed to mitigate 
such error propagation. Table~\ref{tab:global_penalty_scale} compares generation 
quality with and without the global penalty across group sizes 2–5.

\begin{table}[H]
\centering
\caption{\textbf{Comparison of results with / without the global penalty across group sizes.}
Format: \emph{with penalty} / \emph{without penalty}.}
\label{tab:global_penalty_scale}
\small
\setlength{\tabcolsep}{6pt}
\renewcommand{\arraystretch}{1.12}
\begin{tabular}{c|cc|cc|cc}
\hline
$N$ & \multicolumn{2}{c|}{GMR$\downarrow$} & \multicolumn{2}{c|}{GMC$\uparrow$} & \multicolumn{2}{c}{TIF$\downarrow$} \\
\hline
2 & 13.67 & 14.52 & 81.28 & 80.92 & 0.12 & 0.13 \\
3 & 11.63 & 12.44 & 83.74 & 82.12 & 0.12 & 0.13 \\
4 & 12.59 & 14.30 & 82.16 & 81.47 & 0.13 & 0.13 \\
5 & 14.73 & 15.52 & 81.56 & 80.64 & 0.12 & 0.13 \\
\hline
\end{tabular}
\end{table}

Several observations are noteworthy. First, metrics do not degrade monotonically 
with group size—for instance, TIF for 5 dancers (0.12) is lower than for 4 dancers 
(0.13), and GMC for 3 dancers (83.74) outperforms 2 dancers (81.28), suggesting 
the chain formulation benefits from richer conditioning context as more dancers 
are added. Second, the global penalty consistently improves performance across all 
group sizes, confirming that it effectively mitigates error propagation inherent in 
sequential generation rather than simply applying post-hoc corrections.

\subsection{Analysis on Different Optimization Steps and More Number of Dancers}
Table~\ref{tab:otherpenalty} compares generation quality and inference efficiency under different penalty threshold settings. The ChainDance row shows the configuration that achieves the best performance (consistent with the main experiments in the paper). We then examine the trade-offs between generation quality and efficiency across different multiplication factors.

Table~\ref{tab:otherpenalty} shows that fewer optimization steps reduce computation but also weaken the global refinement, leading to higher GMR and lower GMC. As the multiplier increases, the global penalty constraints take effect more strongly, and both GMR and GMC improve, but at the cost of longer inference time. The ChainDance configuration achieves the best overall balance, giving the lowest GMR and the highest GMC while keeping the inference cost within a reasonable range. This confirms that the chosen penalty provides the most stable refinement behavior for multi-dancer generation. We omit TIF metric in this comparison because the variations across settings are minimal.

\begin{table}[b]
\centering
\caption{\textbf{In-depth analysis of the trade-off between inference time and dance quality on the AIOZ-GDance dataset (3 dancers).} Best results are \textbf{bold}; second-best are \underline{underlined}.}
\label{tab:otherpenalty}
\setlength{\tabcolsep}{4.5pt}
\renewcommand{\arraystretch}{1.12}
\small
\begin{tabular}{l|cc|c}
\hline
\multirow{2}{*}{\textbf{Optimization Steps}} &
\multicolumn{2}{c|}{\textbf{Group Dance}} &
\multicolumn{1}{c}{\textbf{Efficiency}}\\
\cline{2-4}
& GMR$\downarrow$ & GMC$\uparrow$ & \makecell{Inf Time\\(min:sec)$\downarrow$} \\
\hline
0$\times$
& 12.01 & \underline{82.68} & \textbf{0:06} \\
0.1$\times$
& 11.83 & 82.04 & \underline{0:08} \\
0.25$\times$
& 11.78 & 81.72 & 0:10\\
0.5$\times$
& \underline{11.52} & 82.35 & 0:11 \\
0.75$\times$
& 11.72 & 82.59 & 0:13 \\
\hline
ChainDance (80 steps)
& \textbf{11.63} & \textbf{83.74} & 0:15\\
\hline
\end{tabular}
\end{table}

Table \ref{tab:moredancers} reports the trade-off between inference time and group-level dance quality as the number of dancers increases. As the group size increases from 2 to 6, inference time increases gradually. Importantly, this growth is accompanied by relatively stable group dance quality, indicating that scalability does not come at the cost of coordination or realism. In particular, GMR and GMC remain competitive across different group sizes, with the best overall balance observed at three dancers, while TIF shows only a mild increase for larger groups, indicating that collision artifacts do not escalate dramatically with scale.

These results suggest that although sequential generation incurs additional inference cost for larger groups, the proposed method maintains consistent coordination and motion realism as the number of dancers increases. This highlights a practical trade-off between scalability and efficiency: higher group sizes naturally require longer generation time, but the quality degradation remains limited and gradual, demonstrating the robustness of the chain-based formulation in multi-dancer scenarios. While not real-time, it is effective for high-quality offline content of arbitrary-sized group motion, whose training data is scarce. Accelerating optimization is future work.

\begin{table}[t]
\centering
\caption{\textbf{In-depth analysis of the trade-off between inference time and dance quality on the AIOZ-GDance dataset for more dancers.} Best results are \textbf{bold}; second-best are \underline{underlined}.}
\label{tab:moredancers}
\setlength{\tabcolsep}{4.5pt}
\renewcommand{\arraystretch}{1.12}
\small
\begin{tabular}{l|ccc|c}
\hline
\multirow{2}{*}{\textbf{Number of Dancers}} &
\multicolumn{3}{c|}{\textbf{Group Dance}} &
\multicolumn{1}{c}{\textbf{Efficiency}}\\
\cline{2-5}
& GMR$\downarrow$ & GMC$\uparrow$ & TIF$\downarrow$ & \makecell{Inf Time\\(min:sec)$\downarrow$} \\
\hline
2
& 13.67 & 81.28 & \underline{0.12} & \textbf{0:07} \\
3
& \textbf{11.63} & \textbf{83.74} & \textbf{0.10} & \underline{0:15} \\
4
& \underline{12.59} & \underline{82.16} & 0.13 & 0:19\\
5
& 14.73 & 81.56 & \underline{0.12} & 0:22 \\
6
& 14.85 & 80.87 & 0.13 & 0:29 \\
\hline
\end{tabular}
\end{table}

\subsection{Quantitative Evidence for Music and Text Controllability}
\label{sec:controllability}

To rigorously validate the responsiveness of ChainDance to its multimodal conditioning signals, we provide quantitative measurements using \textbf{Beat Alignment Score (BAS)} for music alignment and \textbf{R-precision} for text alignment. Since no prior baseline reports such metrics for group dance generation, we report ground-truth (GT) values as upper bounds.

\subsubsection{Matched vs. Mismatched Inputs}
\label{sec:matched_mismatched}

We measure model responsiveness by comparing performance under \emph{matched} inputs (correctly paired music/text and motion) versus \emph{mismatched} inputs (randomly swapped from other categories). A large gap between the two indicates that the model genuinely conditions on the input rather than ignoring it.

\begin{table}[H]
\centering
\caption{\textbf{Quantitative results for music and text alignment.} \emph{Matched} uses correctly aligned inputs; \emph{Mismatched} uses randomly swapped inputs from other categories. A large gap indicates strong responsiveness to the conditioning signal.}
\label{tab:alignment_match}
\footnotesize
\setlength{\tabcolsep}{6pt}
\renewcommand{\arraystretch}{1.15}
\begin{tabular}{lccc}
\toprule
& Matched & Mismatched & Gap$\uparrow$ \\
\midrule
GT (BAS)            & 0.2463 & --     & --     \\
Ours (BAS)          & 0.2247 & 0.0624 & 0.1623 \\
\midrule
GT (R-precision)    & 0.5106 & --     & --     \\
Ours (R-precision)  & 0.4152 & 0.1093 & 0.3059 \\
\bottomrule
\end{tabular}
\end{table}

\noindent As shown in Table~\ref{tab:alignment_match}, ChainDance achieves a substantial gap between matched and mismatched conditions ($+0.1623$ for music, $+0.3059$ for text), providing quantitative evidence that the model is genuinely responsive to both music and text inputs. The matched-condition scores remain close to GT (0.2247 vs. 0.2463 for BAS; 0.4152 vs. 0.5106 for R-precision), indicating that ChainDance generates motions well-aligned with the conditioning signals.

\subsubsection{Genre-wise Statistical Analysis}
\label{sec:genre_wise}

To further demonstrate controllability and generation quality across diverse music genres, we conduct a genre-wise analysis over all 10 music genres in the AIOZ-GDance test set, reporting both alignment metrics (BAS, R-Precision) and group dance quality metrics (GMR, GMC, TIF).

\begin{table}[H]
\centering
\caption{\textbf{Quantitative results across different music genres.} ChainDance maintains stable alignment scores (BAS, R-precision) and consistent group dance quality (GMR, GMC, TIF) across all 10 genres, demonstrating robust dual-condition control.}
\label{tab:genre_wise}
\footnotesize
\setlength{\tabcolsep}{4.5pt}
\renewcommand{\arraystretch}{1.1}
\begin{tabular}{lcccccccc}
\toprule
\multirow{2}{*}{Music genre} &
\multicolumn{2}{c}{BAS} &
\multicolumn{2}{c}{R-precision} &
\multicolumn{3}{c}{Group Dance Quality} \\
\cmidrule(lr){2-3}\cmidrule(lr){4-5}\cmidrule(lr){6-8}
& GT & Ours & GT & Ours & GMR$\downarrow$ & GMC$\uparrow$ & TIF$\downarrow$ \\
\midrule
Disco       & 0.2871 & 0.2661 & 0.5234 & 0.4194 & 13.52 & 83.03 & 0.12 \\
Electronic  & 0.2712 & 0.2562 & 0.5076 & 0.4212 & 13.21 & 82.83 & 0.13 \\
Folk        & 0.2154 & 0.1961 & 0.5112 & 0.4287 & 13.84 & 81.97 & 0.13 \\
Funk        & 0.2176 & 0.1923 & 0.5154 & 0.4431 & 12.87 & 82.42 & 0.12 \\
Indian      & 0.2234 & 0.2001 & 0.5223 & 0.4076 & 13.48 & 83.13 & 0.12 \\
Latin       & 0.2715 & 0.2504 & 0.4989 & 0.3821 & 13.24 & 81.32 & 0.13 \\
Pop         & 0.2203 & 0.1983 & 0.5187 & 0.4102 & 12.76 & 80.98 & 0.12 \\
R\&B        & 0.2728 & 0.2553 & 0.4923 & 0.3954 & 13.49 & 82.84 & 0.13 \\
Rap         & 0.2156 & 0.1929 & 0.5067 & 0.3998 & 11.98 & 83.13 & 0.12 \\
Reggae      & 0.2681 & 0.2393 & 0.5095 & 0.4445 & 12.57 & 82.76 & 0.12 \\
\bottomrule
\end{tabular}
\end{table}

\noindent Table~\ref{tab:genre_wise} shows that BAS scores remain consistently close to GT across all genres (gap typically below 0.03), with no notable degradation in R-precision. Group dance quality metrics (GMR, GMC, TIF) remain stable across all genres, with no genre showing significant performance drops. These results confirm that ChainDance maintains \emph{robust dual-condition control and group coordination across a broad range of musical styles}, rather than being biased toward a particular genre. Together with Table~\ref{tab:alignment_match}, these analyses provide rigorous quantitative evidence supporting the controllability claim made in the main paper.

\subsection{Order Sensitivity Analysis}
\label{sec:order_sensitivity}

A natural concern with sequential generation is whether ChainDance is sensitive to the choice of generation order or leader selection. Although Section~\ref{sec:chaingen} of the main paper shows that chain pairing and random pairing yield comparable results, that experiment evaluates each strategy only once. To rigorously assess the model's sensitivity to generation order, we conduct two additional analyses on the AIOZ-GDance dataset.

We note that the AIOZ-GDance dataset \emph{lacks leader annotations}, and its dancer indexing does not correspond to any physical position. ChainDance therefore does not depend on a predefined order or fixed leader choice—instead, the model is encouraged to focus on \emph{who participates}, \emph{their motion characteristics}, and \emph{their relative spatial relations}, rather than the absolute feature dimension assigned to each dancer.

\subsubsection{All Possible Orders for 3-Dancer Generation}
\label{sec:order_3dancer}

We randomly sample 100 sequences with 3 dancers and evaluate ChainDance under all $3! = 6$ possible generation orders. Results are reported in Table~\ref{tab:order_3dancer}.

\begin{table}[t]
\centering
\caption{\textbf{Evaluation results for 3-dancer generation under all possible generation orders.} The model is evaluated on 100 randomly sampled sequences. Standard deviations across orders are negligible compared to the gap between methods, indicating order-insensitivity.}
\label{tab:order_3dancer}
\footnotesize
\setlength{\tabcolsep}{8pt}
\renewcommand{\arraystretch}{1.1}
\begin{tabular}{cccc}
\toprule
Order & GMR$\downarrow$ & GMC$\uparrow$ & TIF$\downarrow$ \\
\midrule
123 & 11.42 & 83.53 & 0.11 \\
132 & 11.54 & 82.67 & 0.10 \\
213 & 11.14 & 83.76 & 0.10 \\
231 & 11.61 & 83.21 & 0.11 \\
312 & 11.63 & 84.03 & 0.11 \\
321 & 11.73 & 84.52 & 0.11 \\
\midrule
\textbf{Mean $\pm$ Std} & $11.51 \pm 0.22$ & $83.62 \pm 0.67$ & $0.107 \pm 0.005$ \\
\bottomrule
\end{tabular}
\end{table}

\noindent As shown in Table~\ref{tab:order_3dancer}, varying the generation order causes \emph{negligible} performance variance, with standard deviations of only 0.22 (GMR), 0.67 (GMC), and 0.005 (TIF). These variations are far smaller than the gaps between ChainDance and prior baselines (e.g., GMR gap of 3.13 against TCDiff in Table~\ref{tab:groupdance}), indicating that the model is insensitive to the choice of generation order.

This robustness arises from two design choices: (i) the dancer indexing in AIOZ-GDance is not aligned with physical positions, so the model cannot exploit fixed-order priors during training; and (ii) the GCN-based Group-Aware Motion Encoder captures coordination through \emph{relative} positions and connectivity, rather than absolute feature dimensions assigned to each dancer.

\subsubsection{Random Orders for 5-Dancer Generation}
\label{sec:order_5dancer}

To verify that order-insensitivity also holds for larger groups, we further evaluate ChainDance on 5-dancer generation under 4 different random orders.

\begin{table}[H]
\centering
\caption{\textbf{Results for 5-dancer generation under 4 different random orders.} ChainDance maintains stable performance across orders, confirming that order-insensitivity extends to larger group sizes.}
\label{tab:order_5dancer}
\footnotesize
\setlength{\tabcolsep}{8pt}
\renewcommand{\arraystretch}{1.1}
\begin{tabular}{cccc}
\toprule
Run & GMR$\downarrow$ & GMC$\uparrow$ & TIF$\downarrow$ \\
\midrule
1 & 14.87 & 81.21 & 0.13 \\
2 & 14.63 & 81.63 & 0.13 \\
3 & 14.32 & 81.83 & 0.13 \\
4 & 14.21 & 82.02 & 0.12 \\
\midrule
\textbf{Mean $\pm$ Std} & $14.51 \pm 0.31$ & $81.67 \pm 0.35$ & $0.128 \pm 0.005$ \\
\bottomrule
\end{tabular}
\end{table}

\noindent Table~\ref{tab:order_5dancer} confirms that the model remains order-insensitive at larger group sizes, with standard deviations on the same order of magnitude as the 3-dancer case. The observed variations are again much smaller than the performance gaps between competing methods, demonstrating that ChainDance produces \emph{consistent group choreography regardless of the chosen generation order}, even as group size increases.

\section{Limitations and Future Work}
\label{sec:limitations}

While ChainDance addresses group-size rigidity and structurally preserves dancer identity by construction, several limitations remain that point to future research directions.

\paragraph{Inference Latency.}
While our approach avoids the prohibitively expensive $\mathcal{O}((NL)^2)$ training cost associated with joint modeling, its inference latency is relatively high. This is a common bottleneck for diffusion-based models due to the iterative denoising process, which currently limits strictly real-time interactive deployment. However, this latency (measured in seconds rather than minutes) is acceptable for our primary target applications, such as offline animation and video production. Accelerating inference toward real-time interactivity is a natural next step, e.g., via diffusion distillation or consistency-model variants.

\paragraph{Scaling to Massive Crowds.}
While ChainDance addresses flexible group-size modeling, generating highly coordinated motions for extremely large crowds (e.g., $N > 10$) remains a distinct research problem. Although our dynamic growing graph and test-time penalty optimization effectively mitigate spatial inconsistencies for typical group sizes, maintaining uniform global coherence at massive scale becomes increasingly difficult. Our current capacity nonetheless covers the majority of real-world applications, such as generating pop-group (e.g., boy or girl band) choreographies for music videos and gaming, where group sizes rarely exceed ten. To scale seamlessly to massive crowds, future work will explore parallel-friendly mechanisms or hierarchical generation strategies.

\section{More Qualitative Results}
\label{additional_qualitative}

\begin{figure}[H]
    \centering
    \begin{subfigure}[c]{0.4\linewidth}
        \centering
        \includegraphics[width=\linewidth]{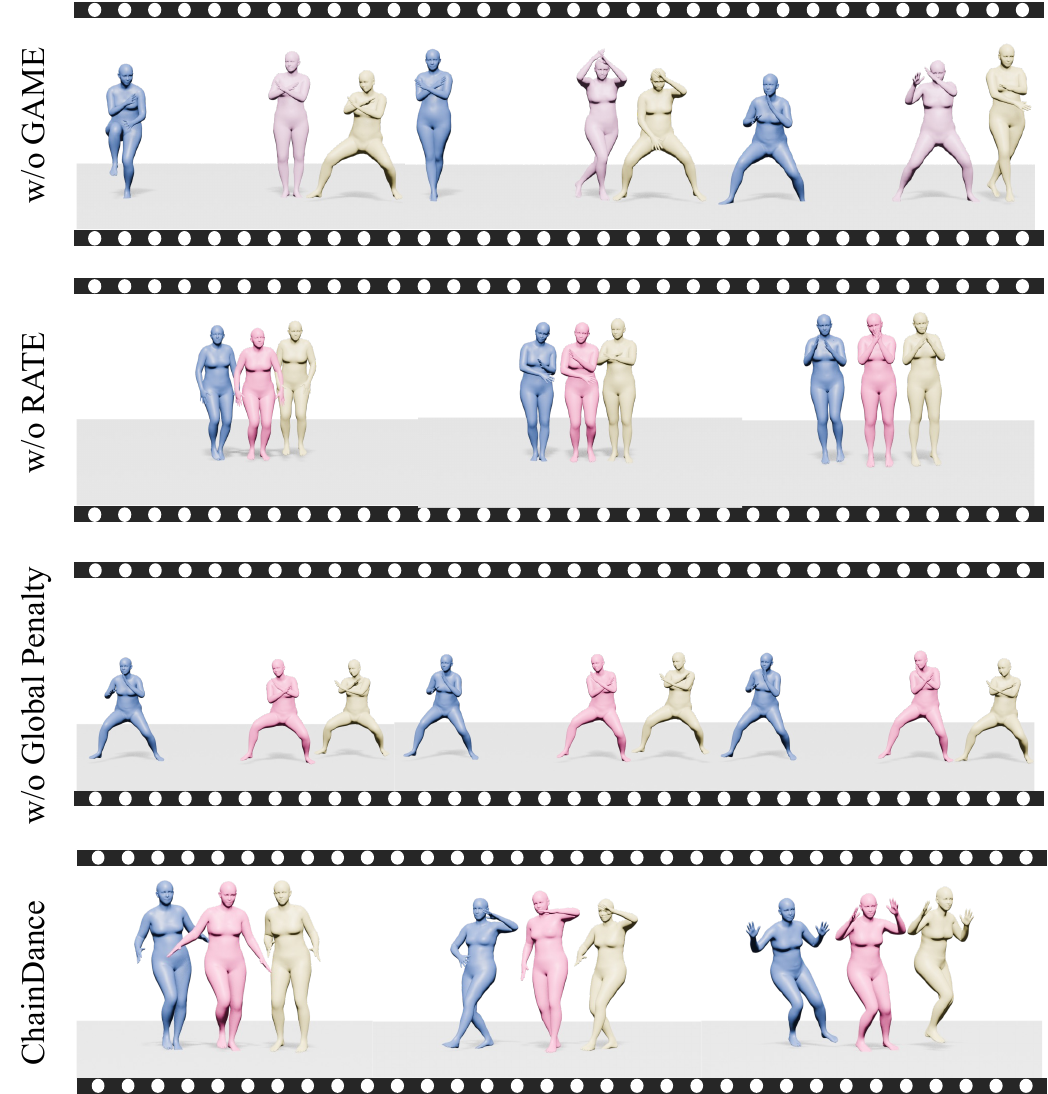}
        \caption{Ablation study}
        \label{fig:viz_abla}
    \end{subfigure}
    \hfill
    \begin{subfigure}[c]{0.55\linewidth}
        \centering
        \includegraphics[width=\linewidth]{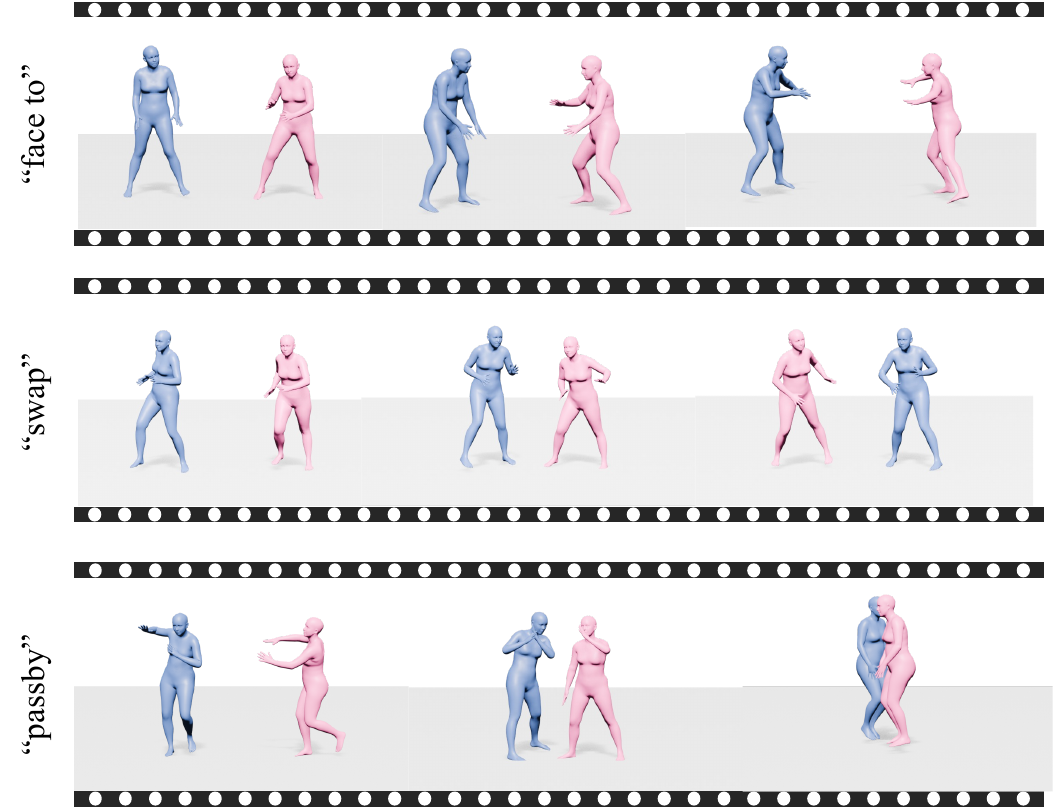}
        \caption{Interaction control via text}
        \label{fig:viz_interact}
    \end{subfigure}
    \caption{\textbf{Qualitative visualizations.} (a) Ablation study showing the effect of each component. (b) Interaction control under different text prompts.}
    \label{fig:viz_combined}
\end{figure}

To visually demonstrate the effect of individual components in ChainDance, we present the qualitative results of the ablation study in Figure~\ref{fig:viz_abla}. Without GAME, each dancer moves almost independently, and the group looks uncoordinated. 
Without text conditioning, the motions of different dancers become too similar and lack variation. 
Without the global penalty, errors accumulate over time, and dancers drift to unreasonable positions with overly large distances. 
These examples show that all three parts are important for coordinated, diverse, and reasonable group choreography.

\begin{figure}[H]
    \centering
    \includegraphics[width=\linewidth]{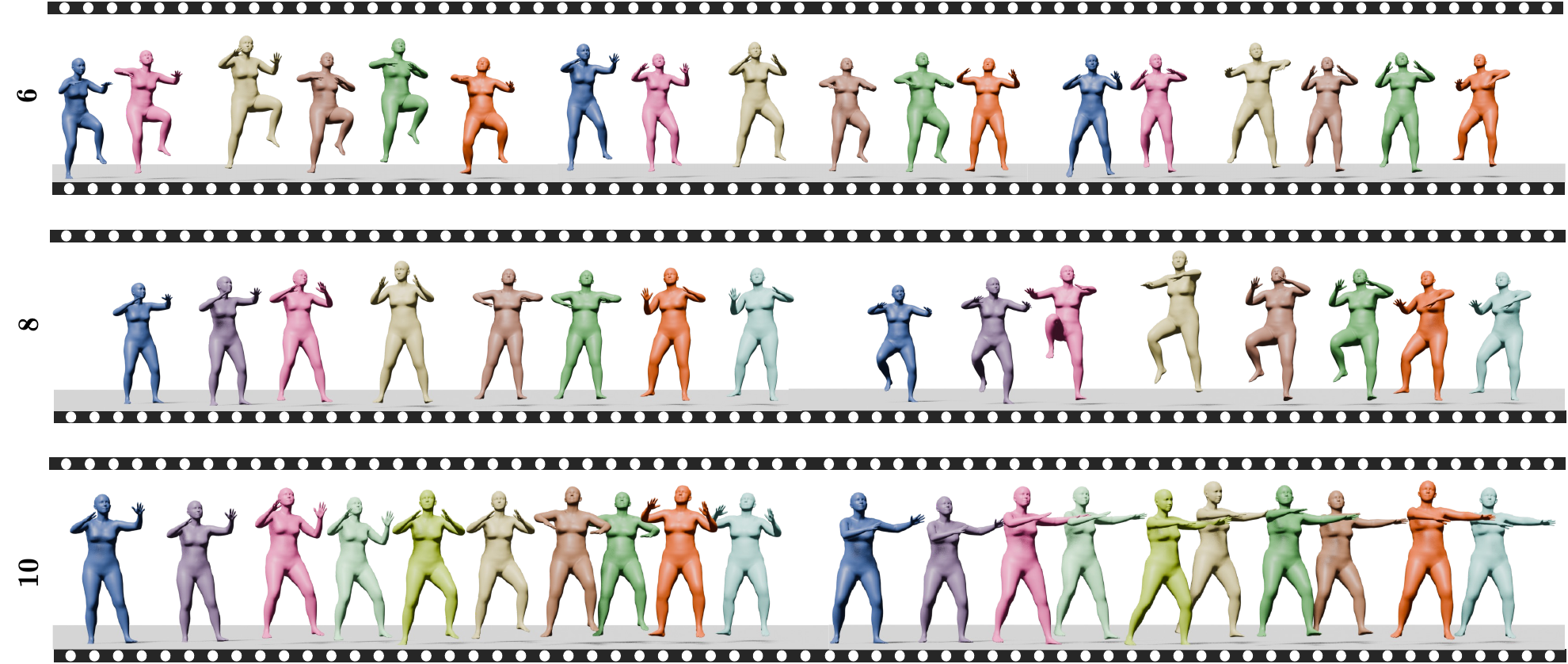}
    \caption{\textbf{ChainDance results on the same music under varying group sizes (from 6 to 10 dancers).} The generated choreography remains coordinated and diverse as the group size scales up.}
    \label{fig:increase}
\end{figure}

\begin{sidewaysfigure*}
    \centering
    \includegraphics[width=\textheight]{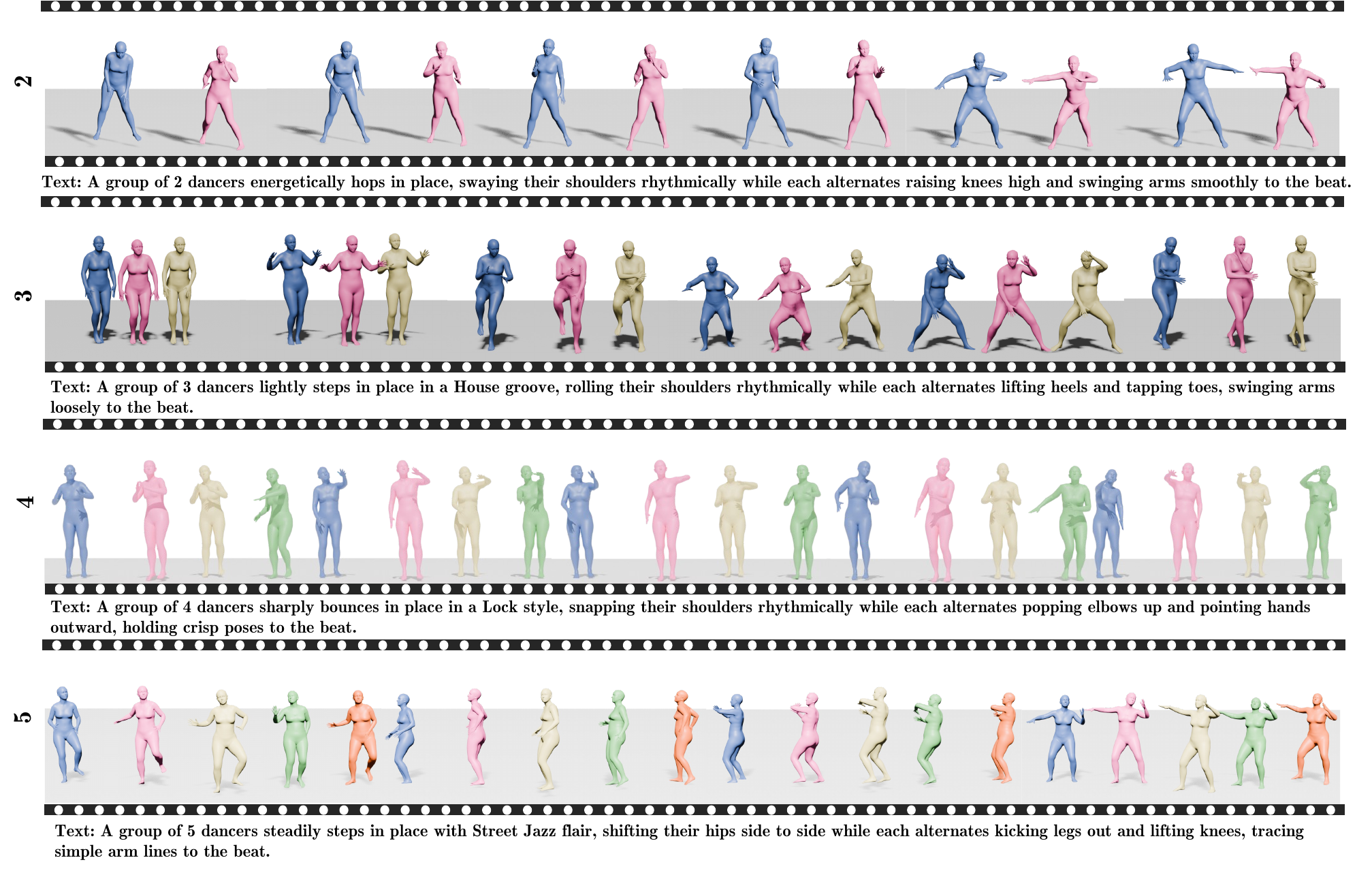}
    \caption{Qualitative results of ChainDance under different group sizes.}
    \label{fig:multi}
\end{sidewaysfigure*}

Then, we provide additional frame-by-frame visualizations of the three types of interaction control in Figure~\ref{fig:viz_interact}, and show ChainDance results under different group sizes in Figure~\ref{fig:multi}. Figure~\ref{fig:increase} further visualizes the process of progressively adding dancers to the same music, showing that ChainDance integrates new dancers smoothly and remains effective for large groups of 6–8 dancers. All examples follow the same visualization settings as in the main paper, where we use the simple SMPL built-in mesh to convert PKL sequences into FBX files. 

For the supplementary video demos included in the \textbf{index.html} file, we further adopt the refined YBot model to render the motions with higher visual quality. The videos are rendered in Blender using scripts adapted from the open-source TCDiff~\cite{dai2025harmonious} code, to which we are grateful.

\end{document}